\documentclass[conference]{IEEEtran}
\IEEEoverridecommandlockouts
\usepackage{spconf}
\usepackage[utf8]{inputenc}
\usepackage{times}
\usepackage{helvet}
\usepackage{courier}
\usepackage{graphicx}

\usepackage[colorlinks=true,linkcolor=black,anchorcolor=black,citecolor=black,filecolor=black,menucolor=black,runcolor=black,urlcolor=black]{hyperref}

\usepackage{url}
\usepackage[T1]{fontenc}    
\usepackage{booktabs}       
\usepackage{amsfonts}       
\usepackage{nicefrac}       
\usepackage{microtype}      
\usepackage[usenames,dvipsnames]{pstricks}
\usepackage{epsfig}
\usepackage{pst-grad} 
\usepackage{pst-plot} 
\usepackage{placeins} 
\usepackage{bbm}
\usepackage{amsmath}
\usepackage{amssymb}
\usepackage{theorem}
\usepackage{mathrsfs} 
\usepackage{euscript}
\usepackage{graphicx}
\usepackage{cases}
\usepackage{algorithm}
\usepackage{algorithmic}

\newcommand{\HH}{\ensuremath{\mathbb{R}^d}}
\newcommand{\LL}{\ensuremath{\mathbb{R}^k}}

\newcommand{\proj}{\mathrm{proj}}

\title{Semi-supervised classification using a supervised autoencoder for biomedical applications}
\name{Cyprien Gille$^{1}$, Frederic Guyard$^{2}$ and Michel Barlaud$^{1}$, Fellow IEEE  }
\address{$^{1}$ Laboratoire I3S CNRS, Cote d'Azur University, Sophia Antipolis, France
\\$^{2}$Orange Labs, Sophia Antipolis, France}

\begin{document}

\maketitle
\begin{abstract}
In this paper we present a new approach to solve semi-supervised classification tasks for biomedical applications, involving a supervised autoencoder network. We create a network architecture that encodes labels into the latent space of an autoencoder, and define a global criterion combining classification and reconstruction losses. We train the Semi-Supervised AutoEncoder (SSAE) on labelled data using a double descent algorithm. 
Then, we classify unlabelled samples using the learned network thanks to a softmax classifier applied to the latent space which provides a classification confidence score for each class.

We implemented our SSAE method using the PyTorch framework for the model, optimizer, schedulers, and loss functions.
We compare our semi-supervised autoencoder method (SSAE) with classical semi-supervised methods such as Label Propagation and Label Spreading, and with a Fully Connected Neural Network (FCNN).
Experiments show that the SSAE outperforms Label Propagation and Spreading and  the Fully Connected Neural Network both on a synthetic dataset and on two real-world biological datasets.
\end{abstract}


\section{Related works}

Annotation of biomedical databases from clinical data by  clinicians is a very difficult, sometimes imprecise, and time consuming task.
An alternative is to ask the clinician expert for the annotations they are the most confident in : We then obtain a semi-supervised classification problem. \\
Semi-supervised learning is a machine learning paradigm \cite{semisupervised} using both labelled and unlabelled data to perform two tasks, classically classification  and clustering. Semi-supervised learning algorithms attempt to improve performance in one of these two tasks by utilizing information generally associated with the other \cite{Semi-SVM,Semi-Zhou,ssl-survey-zhu,Zhu20,supervised_auto,semisup-gauss-harmo}. Subramanya and Talukdar (2014) provided an overview of several graph-based techniques \cite{graph-based-ssl}; Triguero et al. analyzed pseudo-labelling methods \cite{triguero-self-labeled}; Oliver et al.  compared several semi-supervised neural networks, on two image classification problems \cite{oliver-deep-ssl-eval}: Recent research on semi-supervised learning is focused on neural network-based methods (see \cite{Survey-Semi} for a survey). 
The basic idea of semi-supervised learning problems is the prior assumption of consistency using for example spectral methods \cite{zhou2003learning,spectral}\\

In this paper we propose a new approach using Supervised autoencoders (SSAE). Autoencoders were introduced in the field of neural networks decades ago, their most efficient application at the time being dimensionality reduction \cite{deep}.
Autoencoders were also successfully used for data denoising \cite{Autodenoising}, extracting useful and meaningful features.
Meanwhile, Deep generative models have been used to learn generator functions that map points from a low-dimensional latent space to a high-dimensional data space. These generative models include variational autoencoders (VAEs)  \cite{VAE} and generative adversarial networks (GANs)\cite{GAN,GAN2}.
Generative modeling sparked interest in semi-supervised learning contexts, and was used to improve classification performance \cite{VAE,LadderVAE,semisupervised,Rochelle}.\\
Moreover, we propose a constrained regularization approach that takes advantage of a available efficient projection algorithms for the $\ell_1$ constraint \cite{condat}, convex constraints \cite{BBCF}  and structured constraints $\ell_{1,1}$ \cite{BG20}.\\
We combined these two frameworks to propose a new Semi-supervised Autoencoder (SSAE).

We point out the following contributions of our paper :
\begin{itemize}
  \item We create a network architecture that incorporates the ground truth labels into the latent space of an autoencoder.
  \item We define a global criterion combining classification and reconstruction loss.
  \item We provide efficient feature selection of informative features using a structured constraint $\ell_{1,1}$.
  \item  We classify the unlabelled samples using the trained autoencoder network through a softmax classifier applied to the latent space, which transparently offers a prediction confidence score. 
  \item We provide interpretability through the SSAE's informative latent space.
\end{itemize}

\section{Semi-supervised Autoencoder framework}

\subsection{Architecture}
Let us recall that VAE networks encourage their latent space to fit a prior distribution \cite{VAE}, typically a Gaussian. In order to address this simplification, recent papers have proposed latent spaces with more complex distributions (e.g., Gaussian mixtures \cite{mixture}) of the latent vectors, but they are non-adaptive and unfortunately may not match the specific data distribution.

To cope with this issue, our network encourages the latent space to fit a distribution learned from the labels rather than a parametric prior distribution.
Projecting the data in the lower dimension latent space is crucial to be able to separate them accurately.\\
Let us recall that the encoder (or discriminative part) of the autoencoder maps feature points from a high dimensional input space $\HH$ to a low dimensional latent space $\LL$, and that the decoder (or generative part) maps feature points from $\LL $ to $\HH$.

\begin{figure}
    \centering
    \includegraphics[width=.49\textwidth,height=5cm]{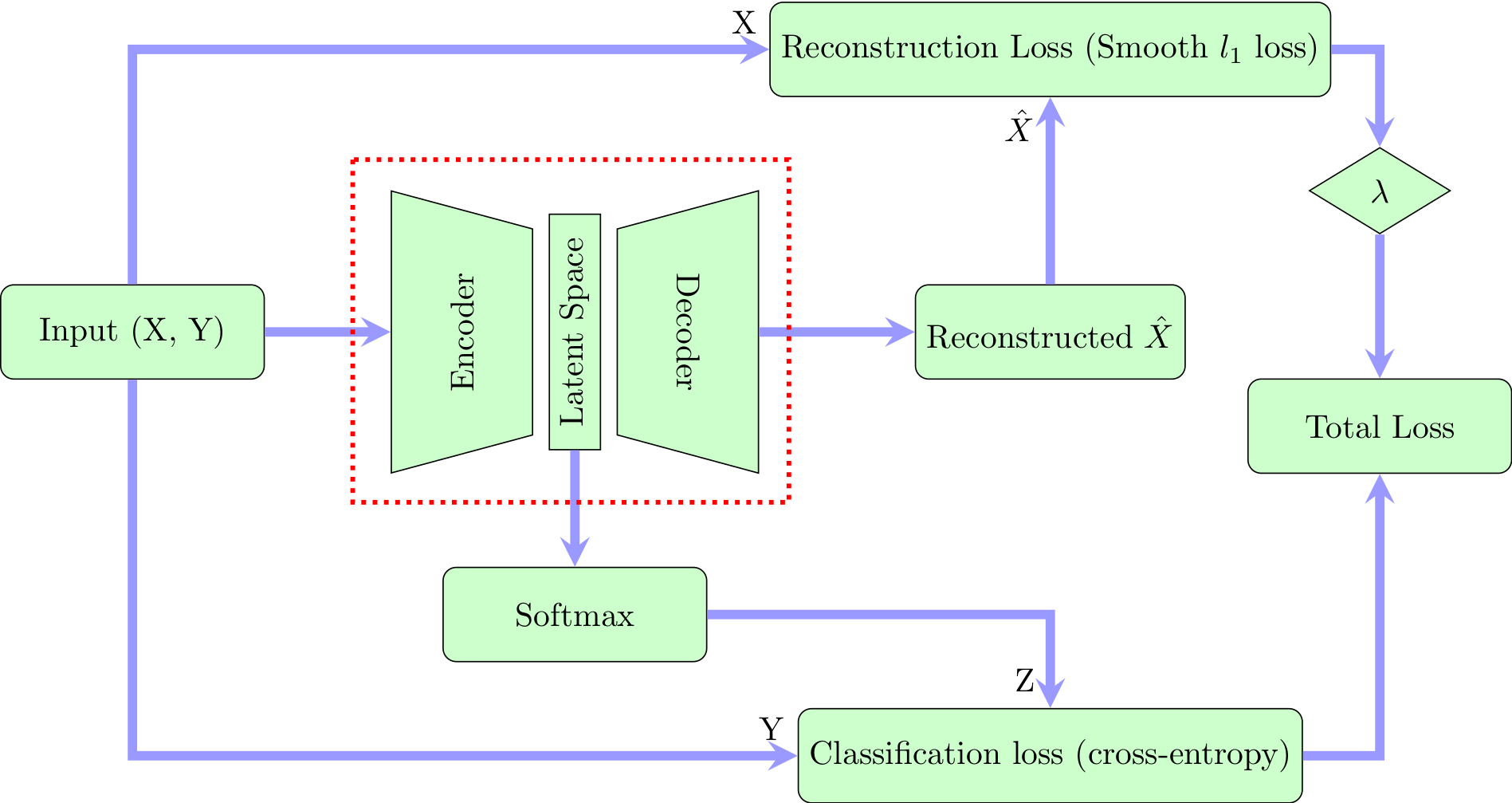}
    \caption{Supervised autoencoder framework}
    \label{NP_auto}
\end{figure}

Figure \ref{NP_auto} depicts the main constituent blocks of
our proposed approach. Note that we added a "soft max" block to our autoencoder to compute the classification loss.\\
Let  $X$ be the dataset in $\HH$, and $Y$ the labels in $\{0, \dots , k\}$, with $k$ the number of classes.
Let Z $\in \LL$ be the encoded latent vectors, $\widehat{X} $ $ \in \HH $ the reconstructed data and W the weights of the neural network. \\ 

\subsection{Criterion}
The goal is to compute the network weights $W$ minimizing the total loss which includes both the classification loss and the reconstruction loss. Hence our strategy for training the SSAE is based on following requirements.

First, we want to classify data in the latent space:
\begin{equation}
Loss(W) =  \mathcal{H} (Z,Y)
\label{classification}
\end{equation}  
where the classification loss $\mathcal{H}$ is a function of the labels and softmaxed latent variable.

Second, we also want to minimize the difference between the reconstructed output and the original input data: 
\begin{equation}
Loss(W) = \psi (\widehat{X}-X)
\end{equation}
where the reconstruction loss $\psi$ is a function of the input and reconstructed output.

Thus we initially propose to minimize the following criterion to train the auto-encoder:
\begin{equation}
\phi =  \mathcal{H}( Z,Y)+ \lambda \psi (\widehat{X} -X) 
\label{criterion}
\end{equation}
where $\lambda$ is a scalar controlling the weight of the reconstruction loss in the global loss.

In an effort to reduce the network's energy and memory footprint, we also want to sensibly sparsify it, following the work proposed in \cite{BG20}, \cite{Lottery}, \cite{Wen2016} and \cite{double}. To do so, instead of the classical computationally expensive lagrangian regularization approach \cite{hrtzER}, we propose to minimize the following constrained approach :
\begin{equation}
Loss(W) =  \mathcal{H}(Z, Y) + \lambda\psi (\widehat{X} -X) \text{ s.t. } \|W\|_1^1 \leq  \eta.
\label{constraint}
\end{equation}

We use the Cross Entropy Loss for the classification loss $\mathcal{H}$. We use the robust Smooth $\ell_1$ (Huber) Loss \cite{Huber} as the reconstruction loss $\psi$.
Note that the dimension of the latent space corresponds to the number of classes, which means that the softmaxed latent vector transparently provides a classification confidence score for each class, which is useful for clinical practice \cite{johnson-beyond}. \\

\subsection{Algorithm}
Let us recall that $\ell_{1,1} $ norm is computed as the maximum $\ell_1$ norm of a column.
We propose the following algorithm: we first compute the radius $t_i$ and then project the rows using the $\ell_1$ adaptive constraint $t_i$ (see \cite{BG20} for more details):\\

Following the work by Frankle and Carbin \cite{Lottery}, further developed by \cite{double} we follow a double descent algorithm, originally proposed as follows: after training a network, set all weights smaller than a given threshold to zero, rewind the rest of the weights to their initial configuration, and then retrain the network from this starting configuration while keeping the zero weights frozen (untrained). 
We train the network using the classical Adam optimizer \cite{Adam}. \\
To achieve structured sparsity, we replace the thresholding by our $\ell_{1,1} $ projection and devise algorithm \ref{algoglobal}.

\begin{algorithm}[!h]
\begin{algorithmic}
\STATE \textit{\# First descent}
\STATE \textbf{Input:} $W_{init},\gamma,\eta$ 
\FOR{$n = 1,\dots,N$}
  \STATE $W \leftarrow A(W,\gamma, \nabla \phi (W) )$
\ENDFOR
\STATE \textit{\# Projection}
  \FOR{$i = 1,\dots,d$} 
  \STATE{$t_i := proj_{\ell_1}((\|v_i\|_1)_{i=1}^l,\eta)$}
  \STATE{$w_i := proj_{\ell_1}(v_i,t_i)$} 
  \ENDFOR
\STATE $(M_0)_{ij} := \mathbbm{1}_{x \ne 0} (w_{ij})$
\STATE \textbf{Output:} $M_0$
\STATE \textit{\# Second descent}
\STATE \textbf{Input:} $W_{init}, M_0, \gamma$
\FOR{$n = 1,\dots,N$}
  \STATE $W \leftarrow A(W,\gamma, \nabla \phi (W,M_0) )$
\ENDFOR
\STATE \textbf{Output:} $W$
\end{algorithmic}
\caption{Double descent algorithm. $\phi$ is the total loss as defined in (\ref{constraint}), $\nabla \phi (W,M_0)$ is the gradient masked by the binary mask $M_0$, $A$ is the Adam optimizer, $N$ is the total number of epochs and $\gamma$ is the learning rate.}
\label{algoglobal}
\end{algorithm}
$\proj_{\ell_1}(V,\eta)$ is the projection on the $\ell_1$-ball of radius $\eta$ which can be computed using fast algorithms \cite{condat,Perez}.
Low values of $\eta$ imply high sparsity of the network. Using the $\ell_{1,1}$ constraint gives us structured sparsity \cite{barlaud2019}, as depicted in figures \ref{MATRIX-SYNTH}, \ref{MATRIX-IPF} and \ref{MATRIX-LUNG} below.

\section{Experimental results}
We implemented our SSAE method using the PyTorch framework for the model, optimizer, schedulers and loss functions.
We used a symmetric linear fully connected network, with the encoder comprised of an input layer of $d$ neurons, one hidden layer followed by a ReLU activation function and a latent layer of dimension $k$.\\
We compare the SSAE with two classical semi-supervised classification techniques based on similarity graphs, Label Spreading (LabSpread) and Label Propagation (LabProp) \cite{labprop}, using their respective implementations in scikit-learn. We also provide comparison with a Fully Connected Neural Network (FCNN) implemented in PyTorch, corresponding to the encoder section of the SSAE.
We evaluated our method on synthetic data and two different biological datasets.
We apply classical log-transform, zero-mean and scaling to the biological datasets. The code to reproduce our results is made available on GitHub\footnote{\url{https://github.com/CyprienGille/Semi-Supervised-AutoEncoder}}

Note that our supervised autoencoder specifically provides several additional features which are especially insightful for biologists, as reported in \cite{bmcSAE}. Namely, only the SSAE provides a two-dimensional latent space where the samples can be visualized, and their respective classifications interpreted. These latent spaces are depicted in figures \ref{LS-SYNTH}, \ref{LS-IPF} and \ref{LS-LUNG} below.

\subsection{Synthetic data}
To generate artificial data to benchmark our SSAE, we use the $make\_classification$ utility from scikit-learn. This generator creates clusters of points that are normally distributed along vertices of a $k$-dimensional hypercube. We are able to control the length of those vertices and thus the separability of the generated dataset.

We generate 1000 points (a number related to the number of points in large biological datasets) with a varying number $d$ of features. We chose $d=1000$  and $d=10,000$ as the two dimensions to test because this is the typical range for biological data, whether it be single-cell or metabolomic. 

\begin{figure}
    \centering
    \includegraphics[width=0.49\textwidth,height=4.cm]{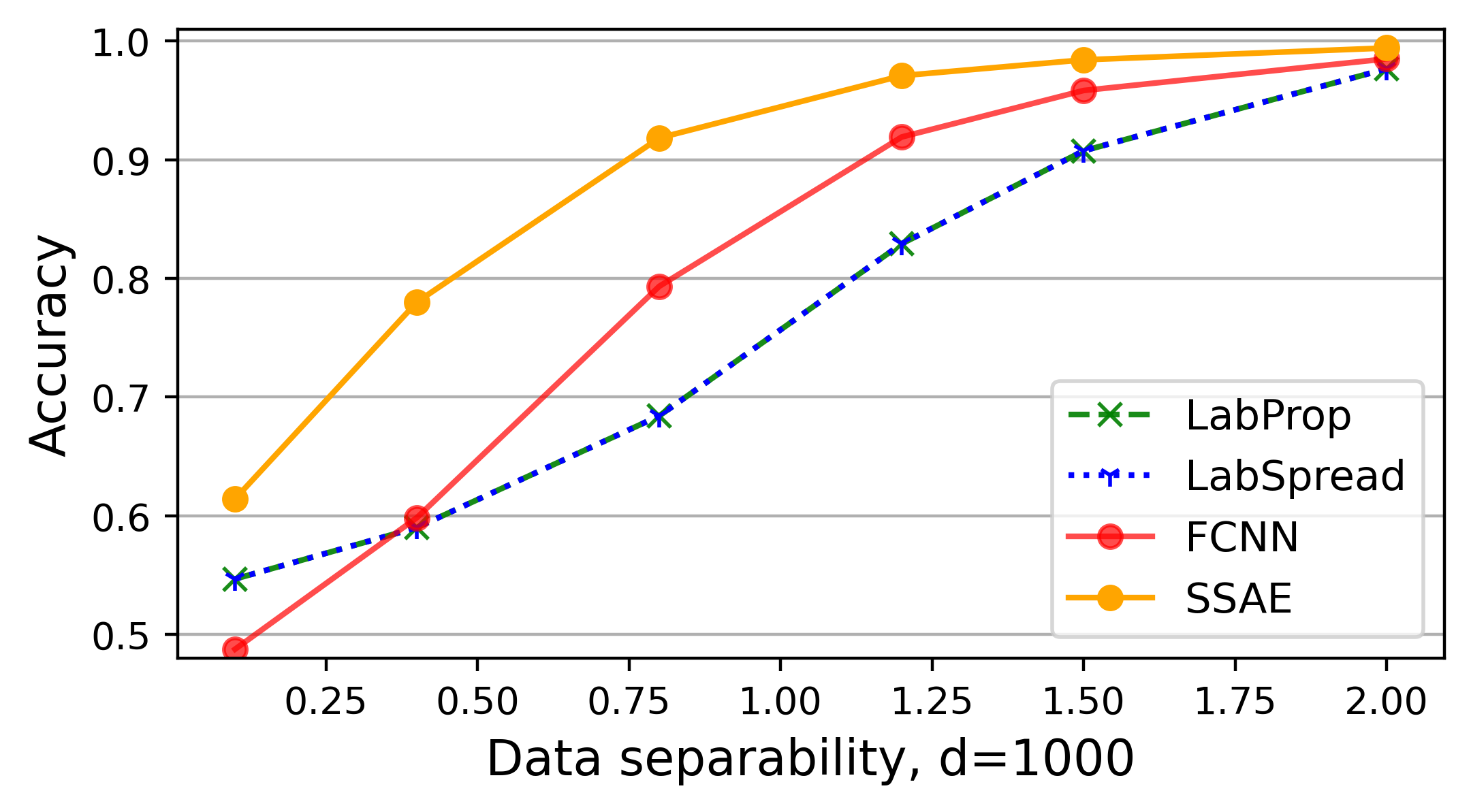}
    \caption{Comparison on synthetic data of our SSAE, the FCNN, Label Propagation and Label Spreading : Accuracy as a function of separability. Mean over 3 seeds, 40\% unlabeled samples, 8 informative features.}
    \label{acc_separ_1k}
\end{figure}

\begin{figure}
    \centering
    \includegraphics[width=0.49\textwidth,height=4.cm]{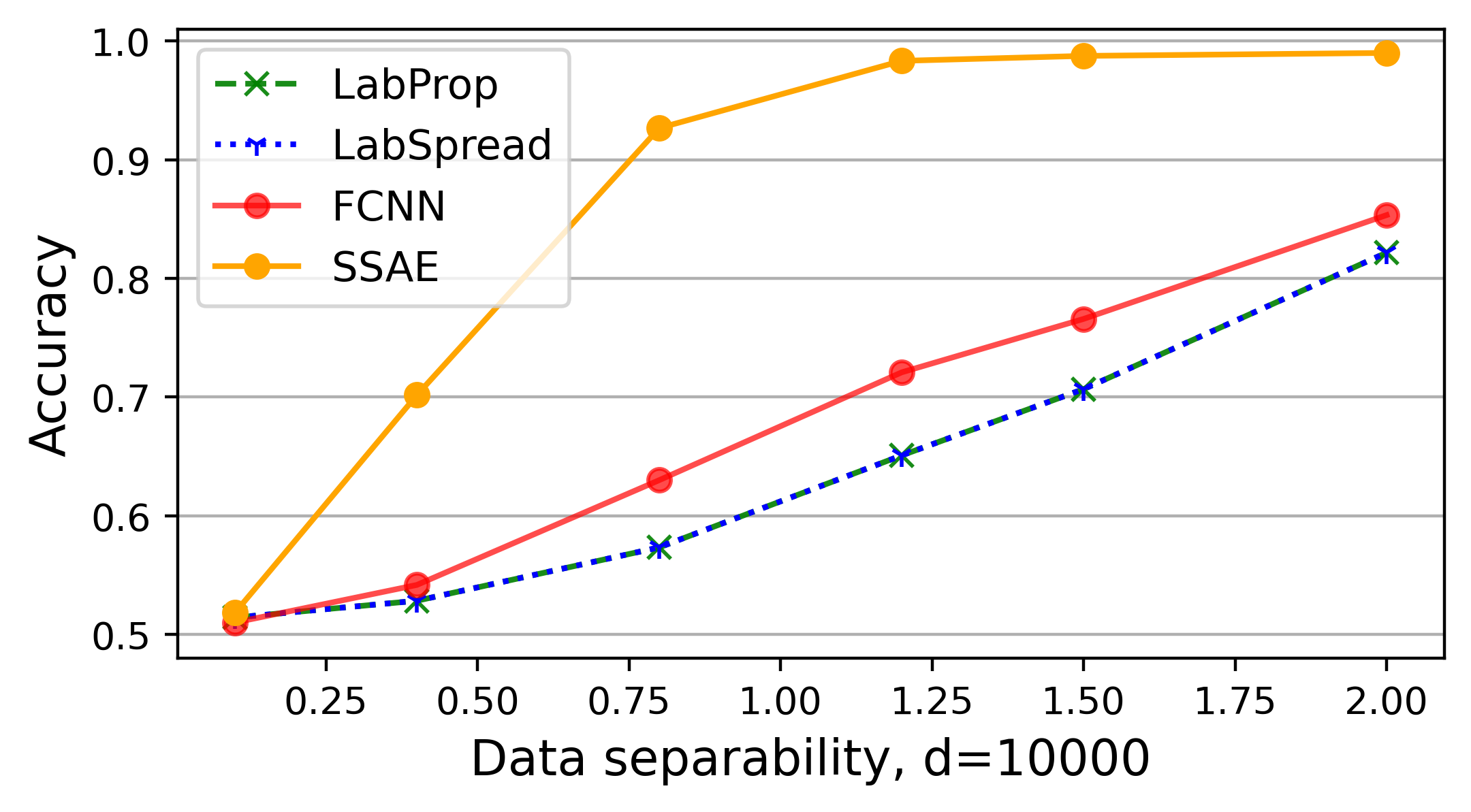}
    \caption{Comparison on synthetic data of our SSAE, the FCNN, Label Propagation and Label Spreading : Accuracy as a function of separability. Mean over 3 seeds, 40\% unlabeled samples, 8 informative features.}
    \label{acc_separ_10k}
\end{figure}

Within this dataset, we randomly pick 40\% of the samples to be considered as unlabelled. We fit or train the algorithms on the remaining samples, and then compute the classification accuracy by comparing the labels predicted by the learned network to the original labels. We report the results in figures \ref{acc_separ_1k} and \ref{acc_separ_10k}.


\begin{figure}
    \centering
    \includegraphics[width=0.49\textwidth,height=4.5cm]{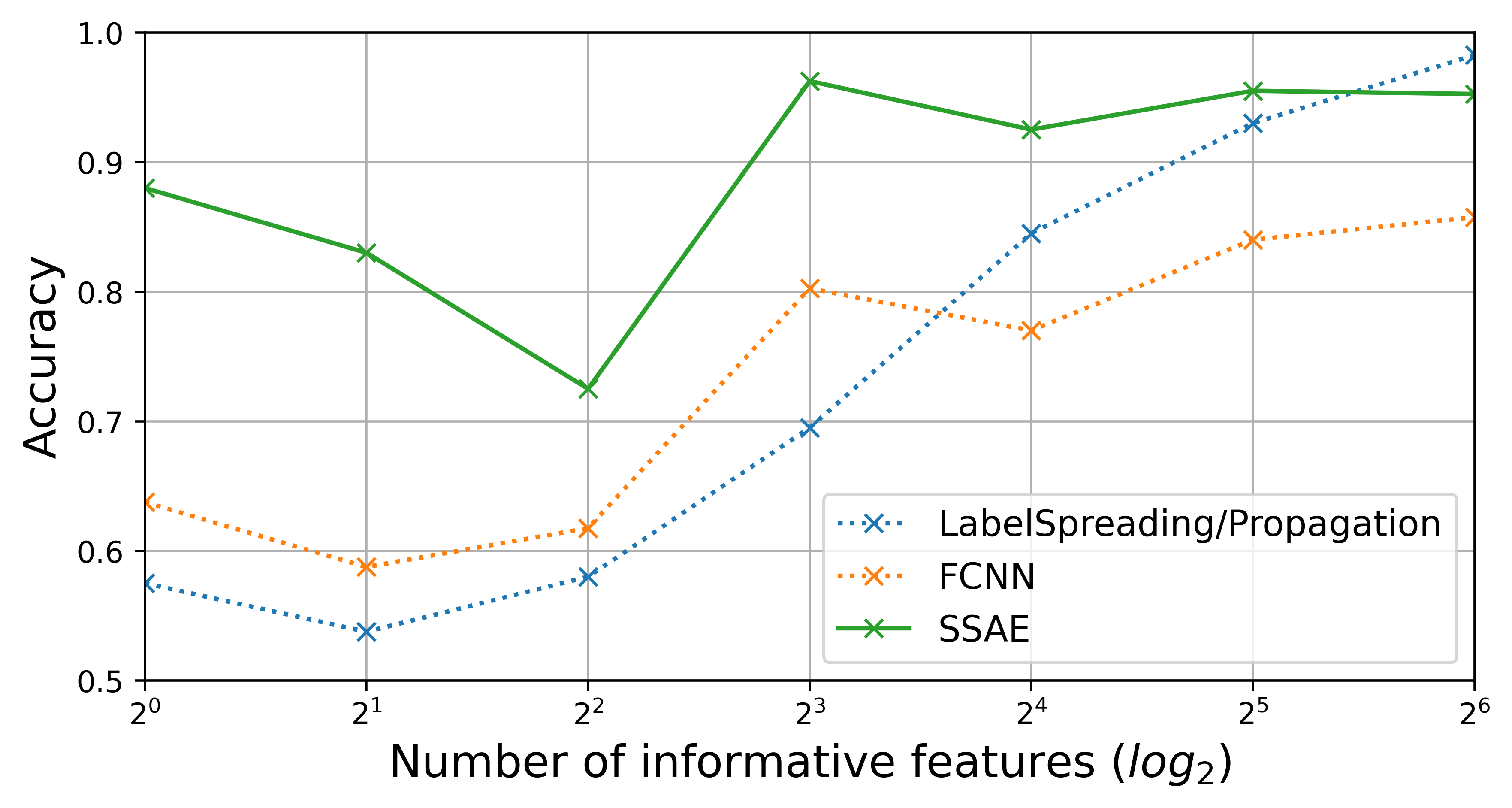}
    \includegraphics[width=0.49\textwidth,height=4.5cm]{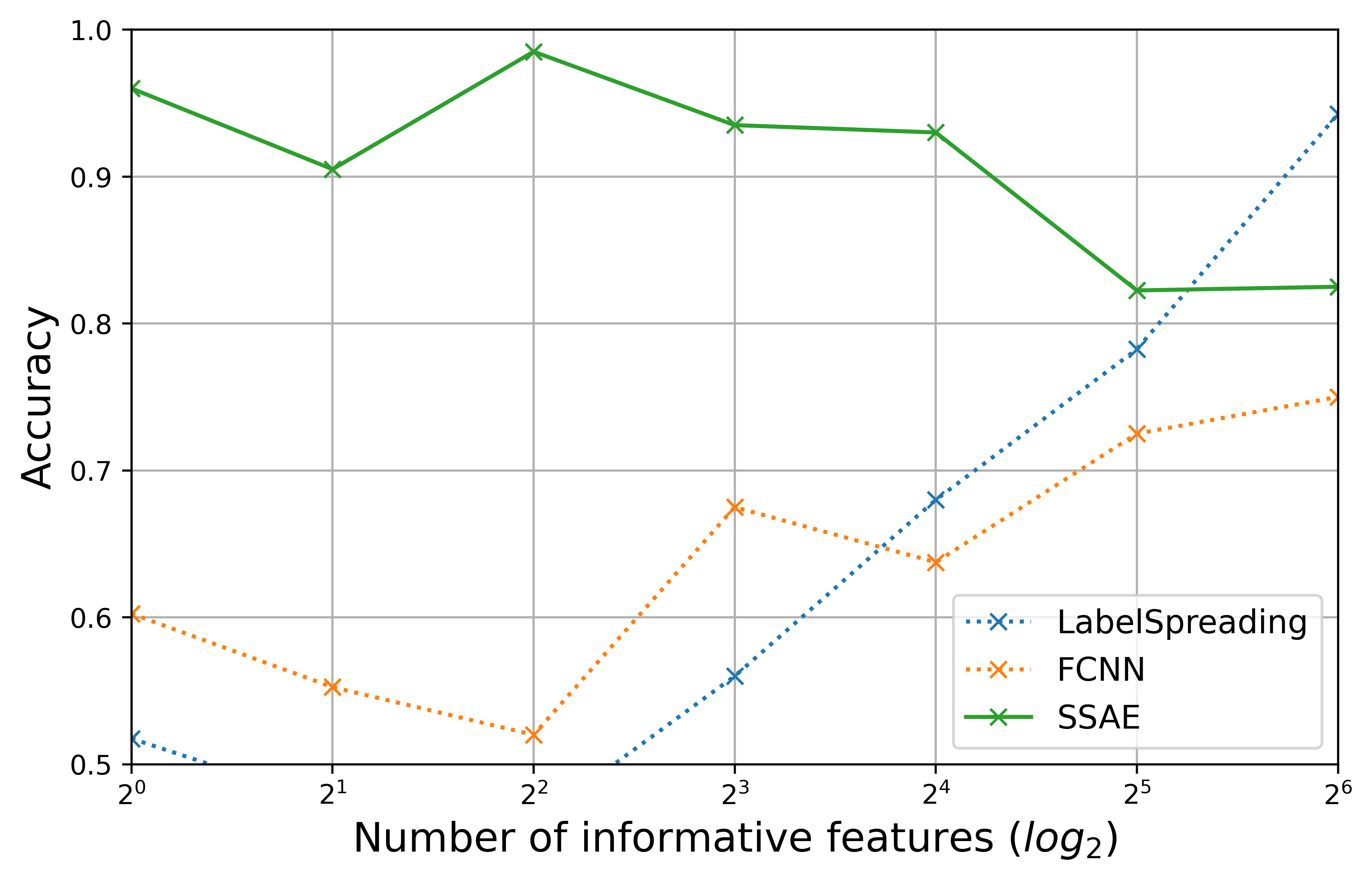}
    \caption{Comparison on synthetic data of our SSAE, the FCNN, Label Propagation and Label Spreading : Accuracy as a function of the number of informative features. 40\% unlabeled samples, 8 informative features. Top: $d=1000$, Bottom: $d=10,000$}
    \label{acc_informative}
\end{figure}

\begin{table}
    \centering
    
    \begin{tabular}{|c|c|c|c|c|c|c|c|}
        \hline
        Lung            & SSAE           & LProp  & LSpread & FCNN  \\
        \hline
        Accuracy $ \%$  & 96.75          &  69.50 &  69.50  & 85.00\\
        \hline
        AUC             &   0.9895       & 0.7871  &   0.7811   & 0.9180\\
       \hline
        F1 score        &   0.9675       & 0.6948 & 0.6948  & 0.8499\\
        \hline
    \end{tabular}
    \caption{Synthetic dataset: comparison of LabelPropagation, LabelSpreading, FCNN and SSAE. 40\% of unlabeled data, separability of 0.8, d= 1000, 8 informative features.}
    \label{SYNTH-TABLE-1k}
\end{table}

\begin{table}
    \centering
    
    \begin{tabular}{|c|c|c|c|c|c|c|c|}
        \hline
        Lung            & SSAE           & LProp  & LSpread & FCNN  \\
        \hline
        Accuracy $ \%$  & 96.00          &  56.00 &  56.00  & 67.50\\
        \hline
        AUC             &   0.9944       & 0.5658  &   0.5635   & 0.7313\\
       \hline
        F1 score        &   0.9599       & 0.5584 & 0.55584  & 0.6749\\
        \hline
    \end{tabular}
    \caption{Synthetic dataset : comparison of LabelPropagation, LabelSpreading, FCNN and SSAE. 40\% of unlabeled data, separability of 0.8, d= 10,000, 8 informative features.}
    \label{SYNTH-TABLE-10k}
\end{table}

Figures \ref{acc_separ_1k} and \ref{acc_separ_10k} shows that our SSAE largely outperforms the classical methods, and by a widening margin when the number of features increases. Classical methods such as label propagation and label spreading are based on a similarity matrix and thus suffer from the curse of dimensionality (the similarity function used in this paper is a $kNN$ algorithm, which has to compute the distance between two samples). As the dimension increases, vectors become indiscernible \cite{curse,hub} and the predictive power of the aforementioned methods is substantially reduced.
Notably, figure \ref{acc_separ_10k} also demonstrates that in higher dimensions, the performance of the FCNN falls off in comparison to that of the SSAE.

Tables \ref{SYNTH-TABLE-1k} and \ref{SYNTH-TABLE-10k} confirm the previous results: the SSAE outperforms the two classical methods across all metrics. More notably, our SSAE also outperforms the F1 score of the FCNN by $12\%$ for $d=1000$, and by $29\%$ for $d=10,000$.

\begin{figure}
    \centering
    \includegraphics[width=0.49\textwidth,height=3.cm]{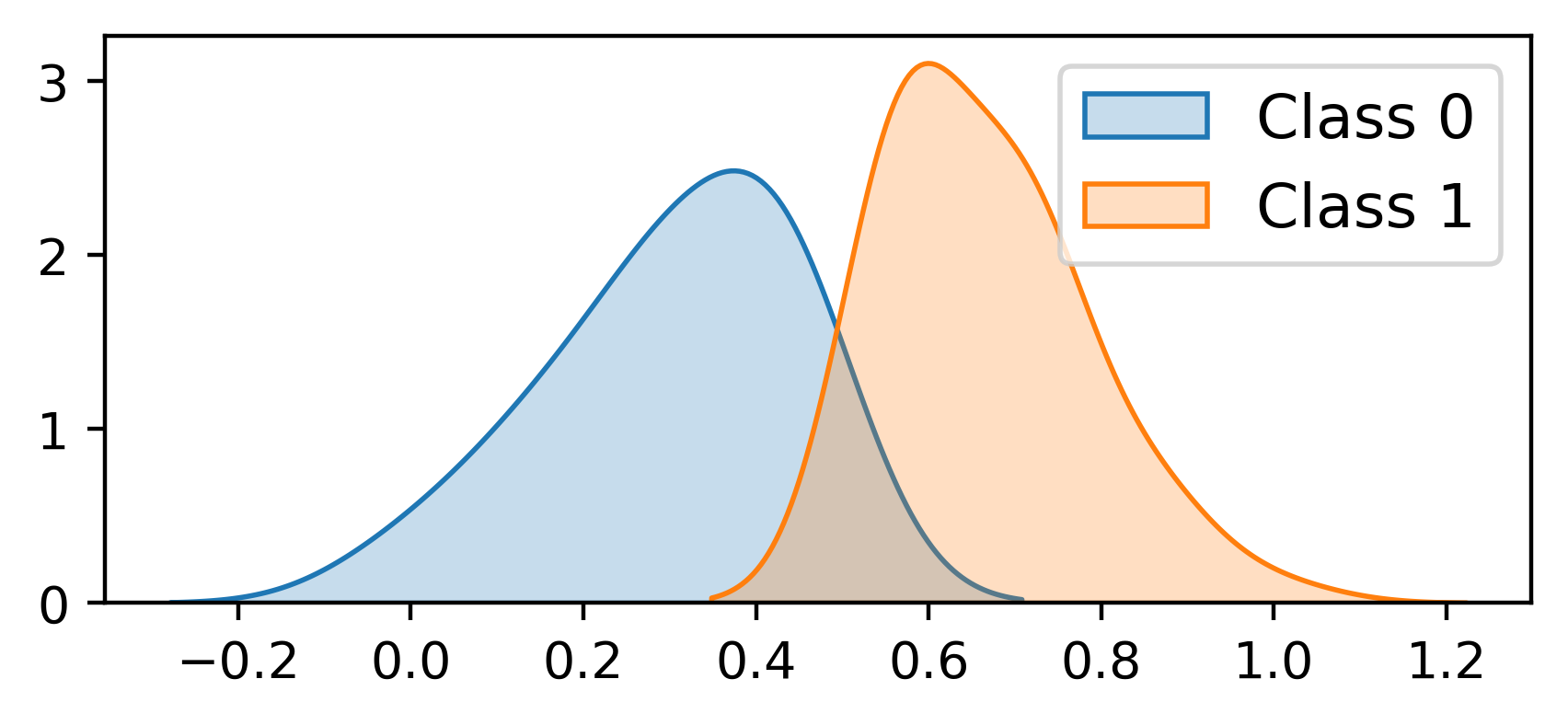}
    \includegraphics[width=0.49\textwidth,height=3.cm]{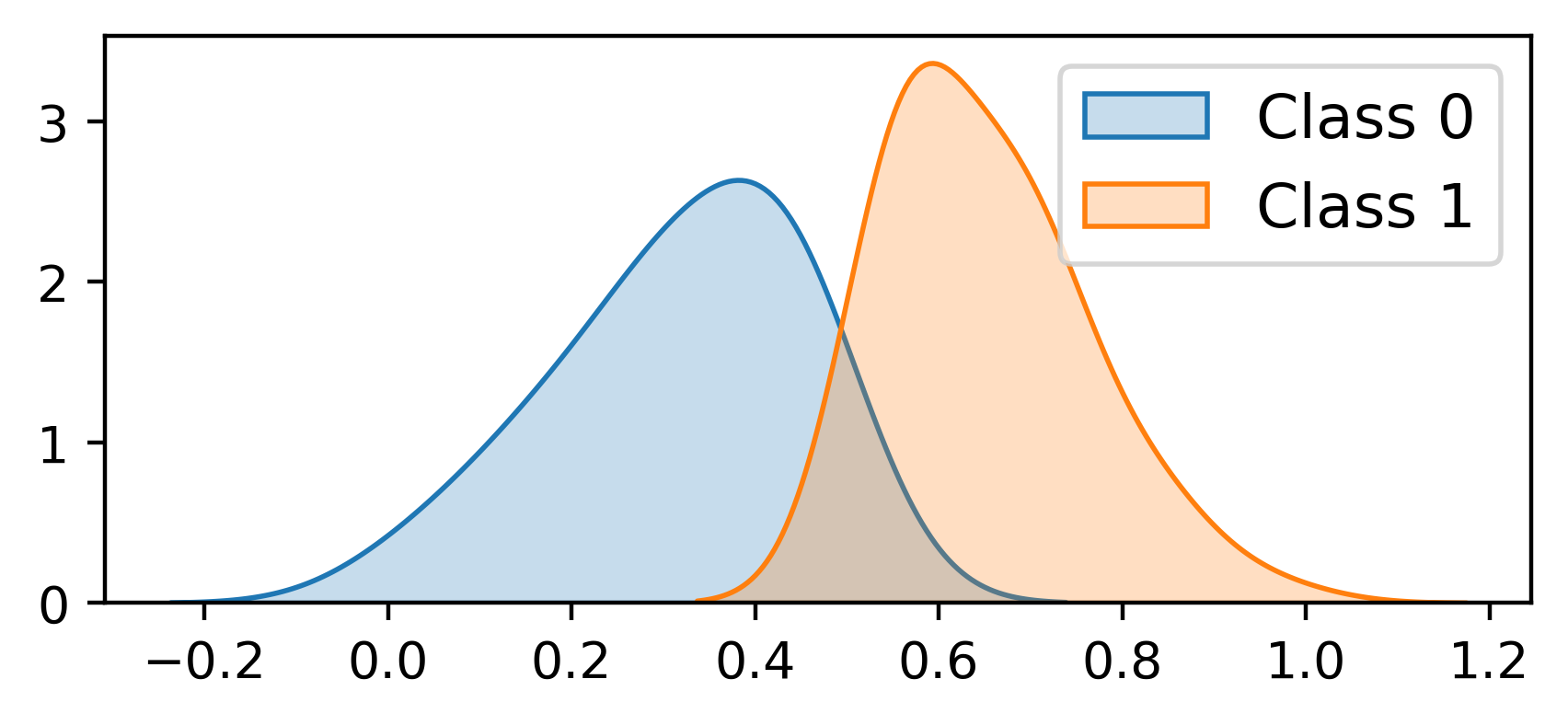}
    \includegraphics[width=0.49\textwidth,height=3.cm]{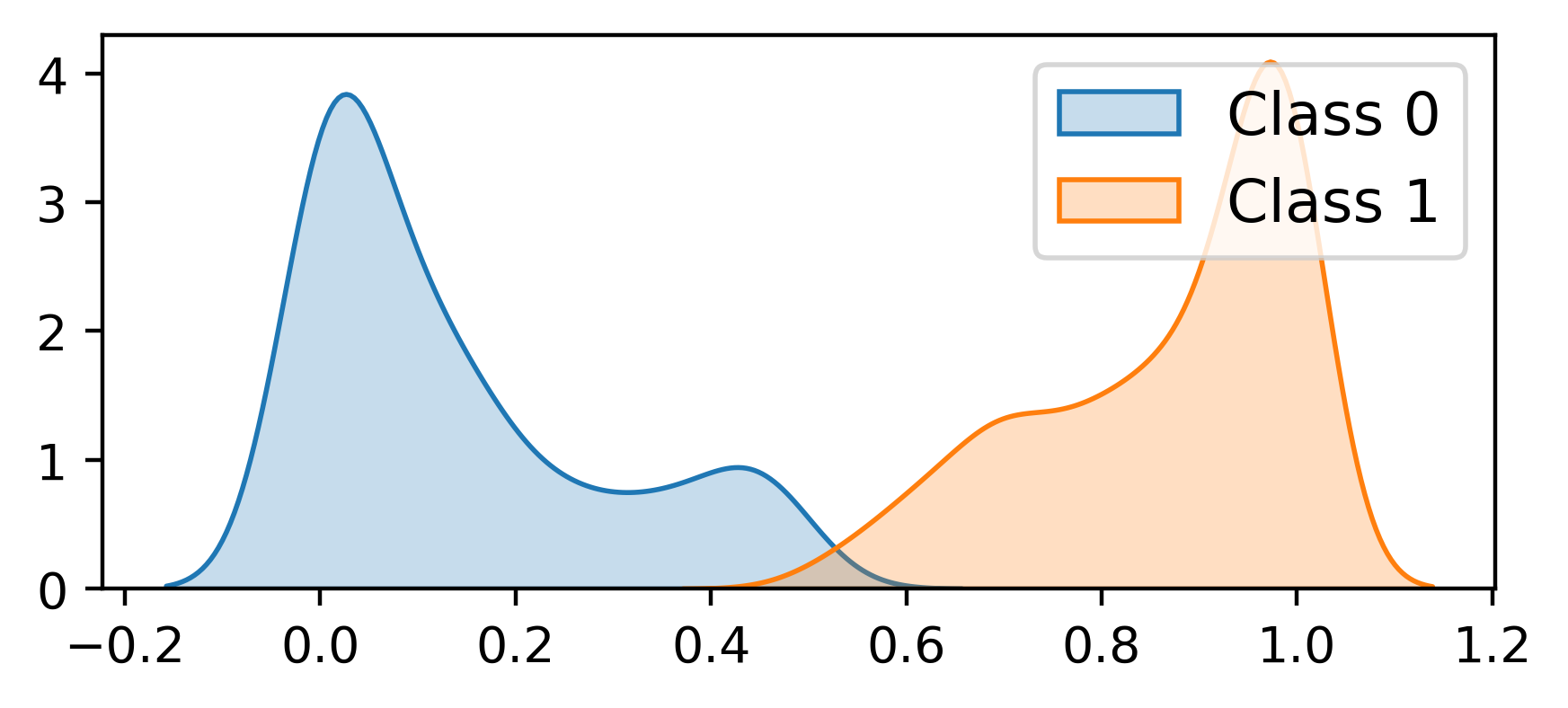}
    \includegraphics[width=0.49\textwidth,height=3.cm]{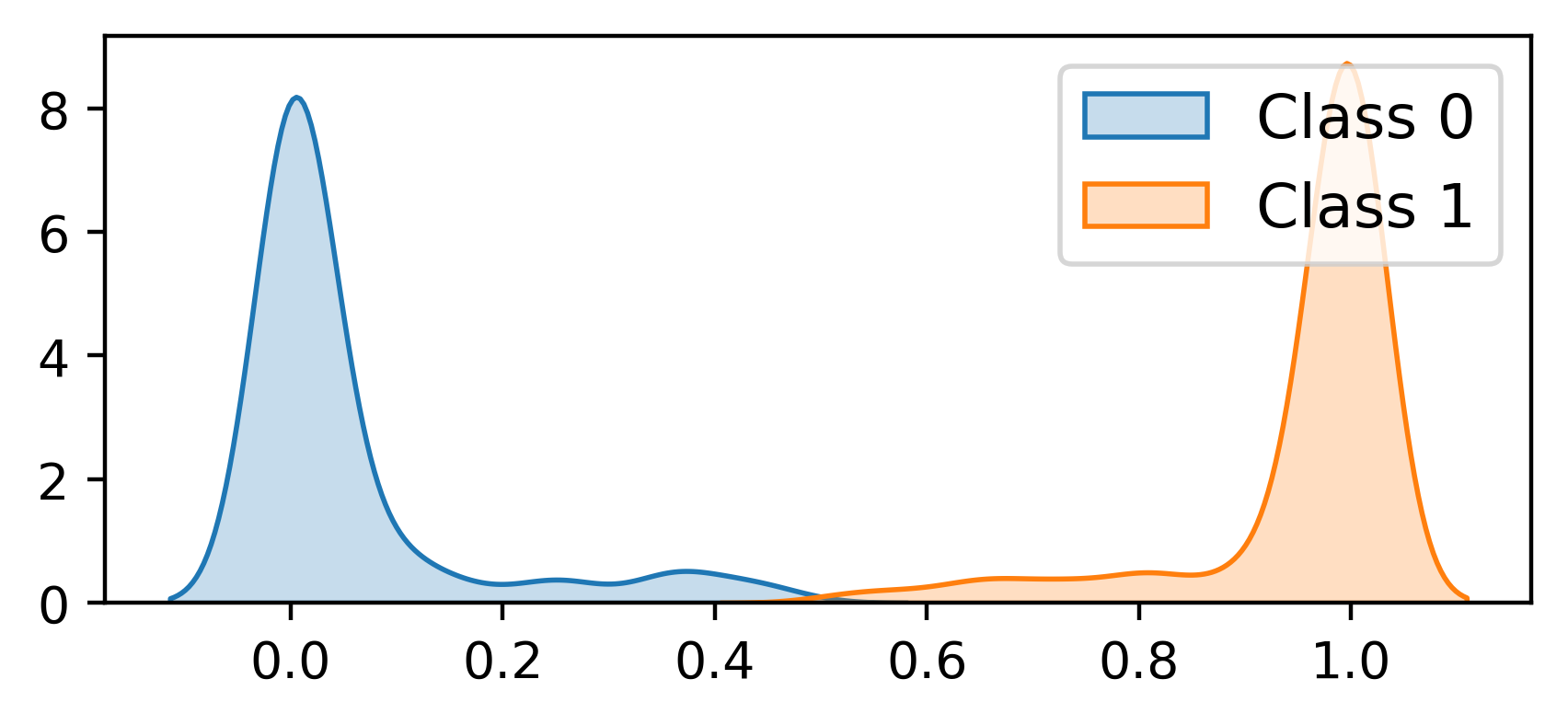}
    \caption{Synthetic dataset, $d=1000$, separability of $0.8$, unlabeled proportion of 40\%, 8 informative features. Comparison of the prediction score distributions. From top to bottom : LabelPropagation, LabelSpreading, FCNN, SSAE with $\ell_{1,1}$}
    \label{distrib-synth}
\end{figure}

Figure \ref{distrib-synth} shows the distribution of the scores of the predicted labels: the two classical methods show poor decision capabilities, as the score of the predicted class will often be close to 0.5. On the other hand, the neural networks are more confident about their predictions, with the SSAE being a better discriminator than the FCNN, which is reflected in their respective metrics in table \ref{SYNTH-TABLE-10k}. Note that the distributions are smoothed using a gaussian kernel, and that the class $0$ distribution is flipped around $0.5$ to better highlight the separation of the labels.

\begin{figure}
    \centering
    \includegraphics[trim={0 0 0 0.8cm},clip,width=0.49\textwidth,height=4.cm]{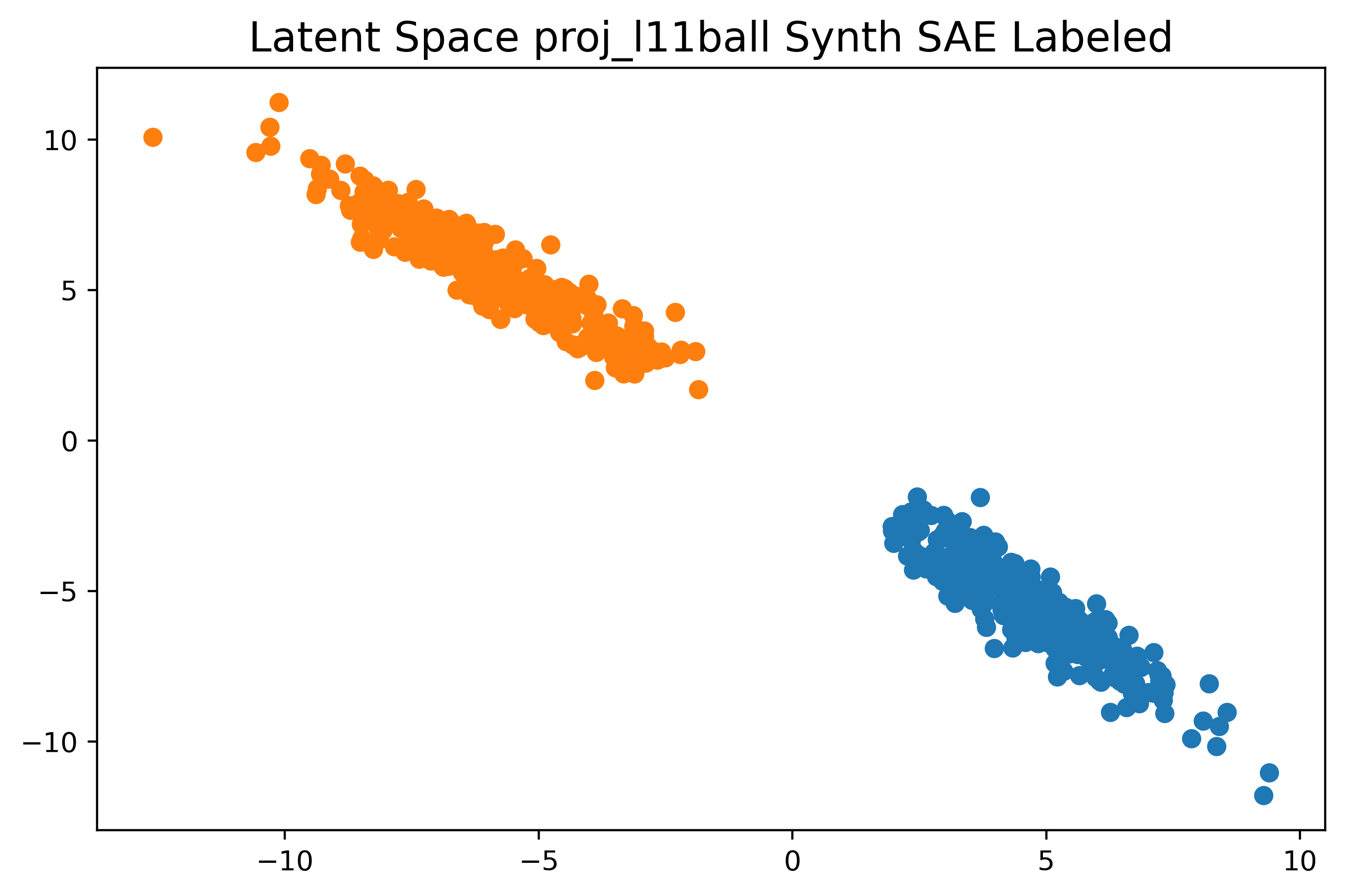}
    \includegraphics[trim={0 0 0 0.8cm},clip,width=0.49\textwidth,height=4.cm]{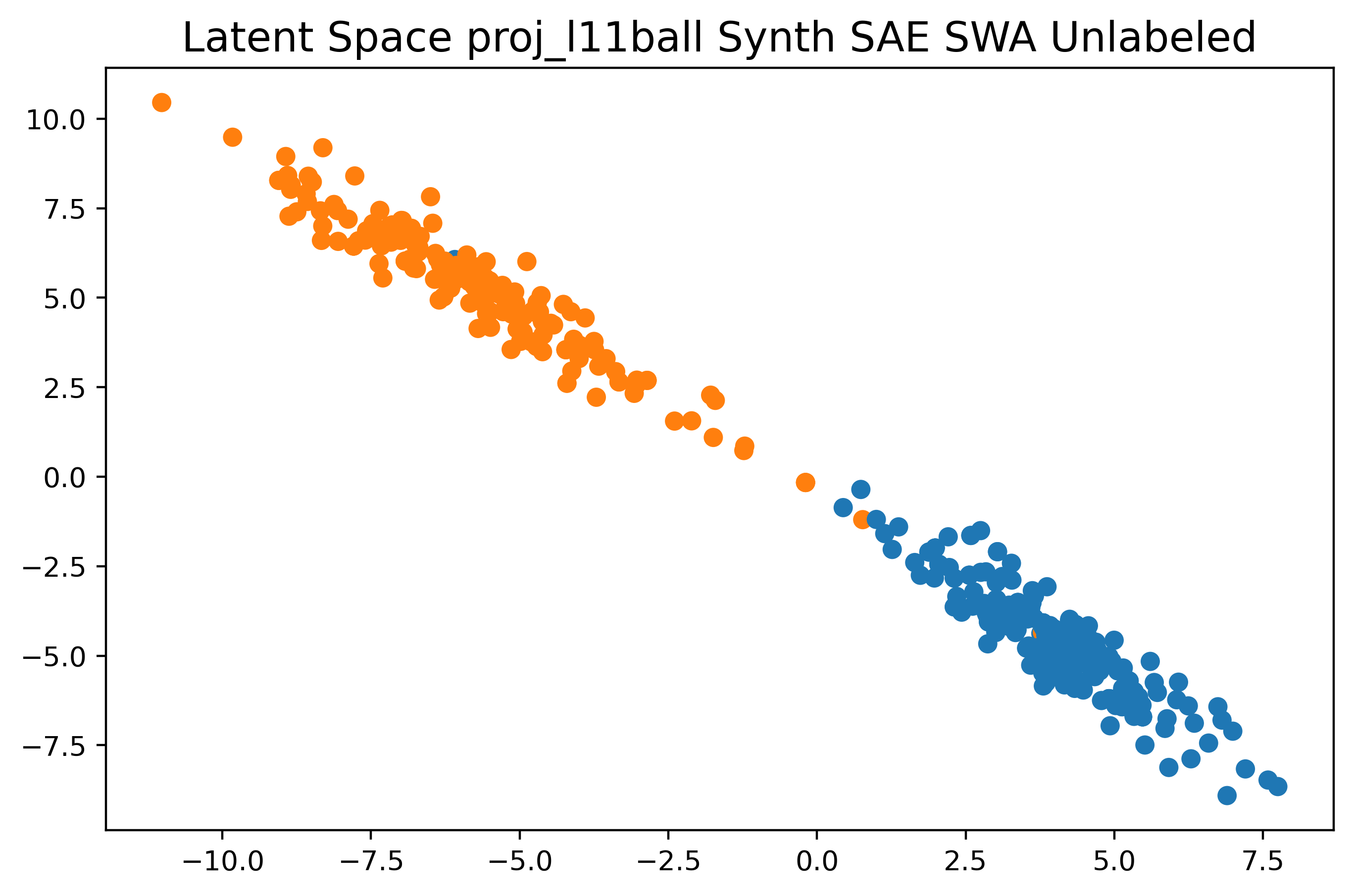}
    \caption{Synthethic dataset, $d=1000$, 40\% of unlabeled samples, Separability of $0.8$, 8 informative features. Samples represented in the latent space of the SSAE. Top: Labeled samples, Bottom: Unlabeled samples.}
    \label{LS-SYNTH}
\end{figure}

Figure \ref{LS-SYNTH} shows the latent space of the SSAE. We can see that, after learning, the SSAE is able to accurately separate the labeled samples, and that the unlabeled samples have been relatively clustered according to their labels. This feature, offered only by the SSAE, provides interpretability to the results and an insightful tool for practical use. A new patient's position in the latent space, as well as the influence of a given input feature on it, can grant acumen to the practician.

\begin{figure}
    \centering
    \includegraphics[trim={0 0 0 0.8cm},clip,width=0.49\textwidth,height=4.cm]{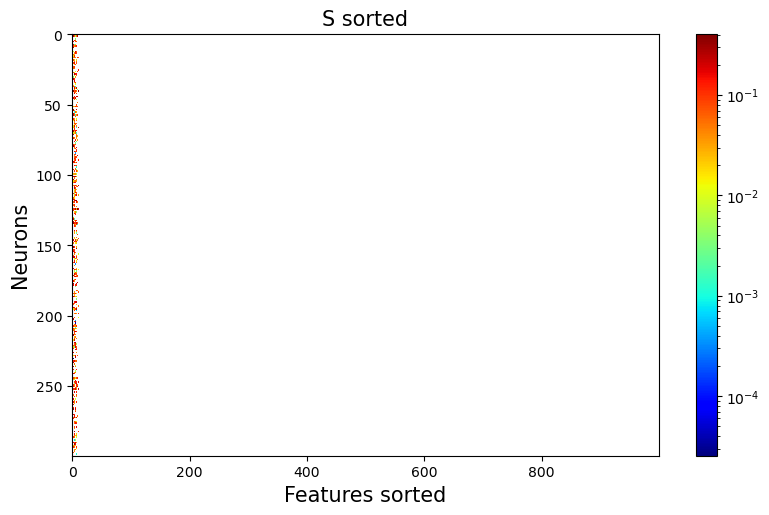}
    \caption{Synthethic dataset, $d=1000$, 40\% of unlabeled samples, 8 informative features. Weight matrix after projection sorted by feature importance. Logarithmic color scale, zero weights are colored in white.}
    \label{MATRIX-SYNTH}
\end{figure}

Figure \ref{MATRIX-SYNTH} shows the activation of each neuron by each input feature, with the features sorted as to have the one of highest norm on the far left. We can see the effect of the projection described in algorithm \ref{algoglobal} : the SSAE performs feature selection and presents substantial sparsity.

\subsection{Biological datasets}
The \textbf{IPF} dataset is a single cell RNA seq \cite{SingleCell} published database which is made up of human fibroblasts transcriptomic profiles, obtained from lung explants of patients with Idiopathic Pulmonary Fibrosis and from healthy donors.
This dataset comes from a study \cite{ipfcellatlas} aimed at characterizing the transcriptional changes induced by the pathology in pulmonary cell types.
The $1443$ samples are described by $14369$ numerical features with high sparsity. From this labelled dataset, we randomly pick a subset of samples to be considered unlabelled, following the same procedure as described in section III.A.

\begin{table}
    \centering
    
    \begin{tabular}{|c|c|c|c|c|c|c|c|}
        \hline
        IPF             & SSAE      & LProp  & LSpread & FCNN        \\
        \hline
        Accuracy $ \%$  & 96.66     & 72.14  & 72.14   & 95.55     \\
        \hline
        AUC             & 0.9947    & 0.7792 &  0.7730    & 0.9903    \\
       \hline
        F1 score        & 0.9633    & 0.6977 &  0.6976 & 0.9510    \\
        \hline
    \end{tabular}
    \caption{\textbf{IPF} dataset: Mean Metrics over 3 seeds : comparison of LabelPropagation, LabelSpreading, FCNN and SSAE. 40\% of unlabeled data.}
    \label{IPF-TABLE}
\end{table}

Table \ref{IPF-TABLE} shows that for the \textbf{IPF} dataset (which is a high-dimensional dataset) only the neural networks manage to accurately classify the unlabelled samples; they both do so almost optimally, reaching very high metrics.\\

\begin{figure}
    \centering
    \includegraphics[width=0.49\textwidth,height=3.5cm]{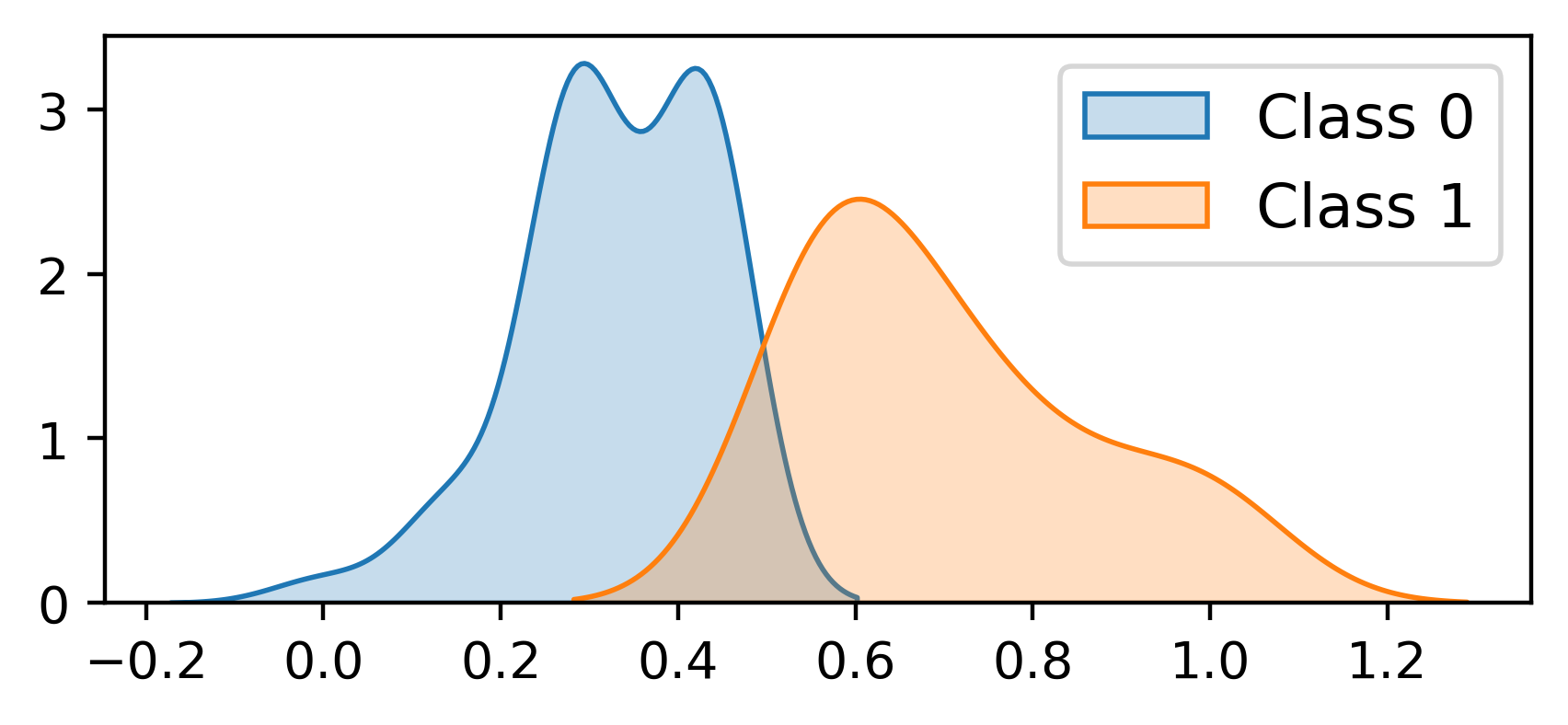}
    \includegraphics[width=0.49\textwidth,height=3.5cm]{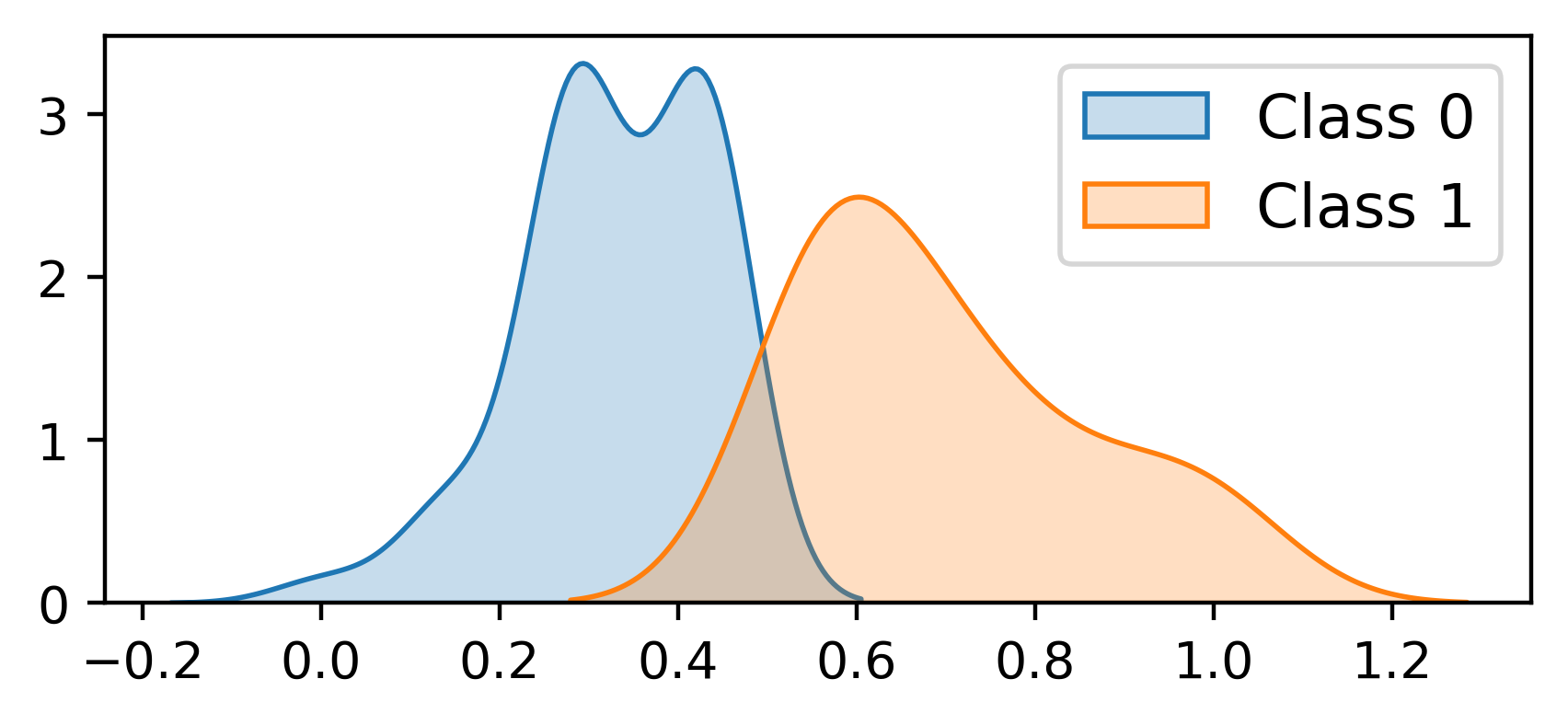}
    \includegraphics[width=0.49\textwidth,height=3.5cm]{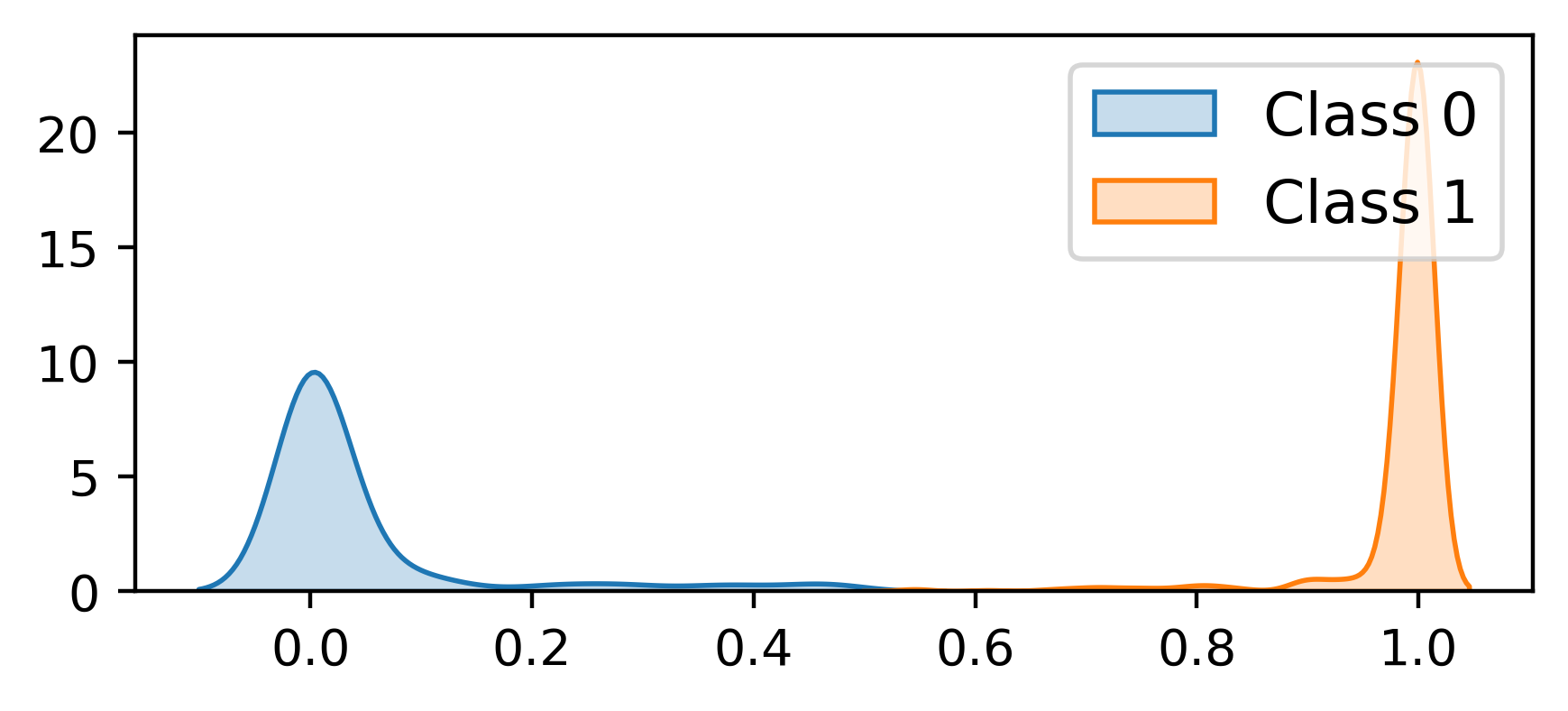}
    \includegraphics[width=0.49\textwidth,height=3.5cm]{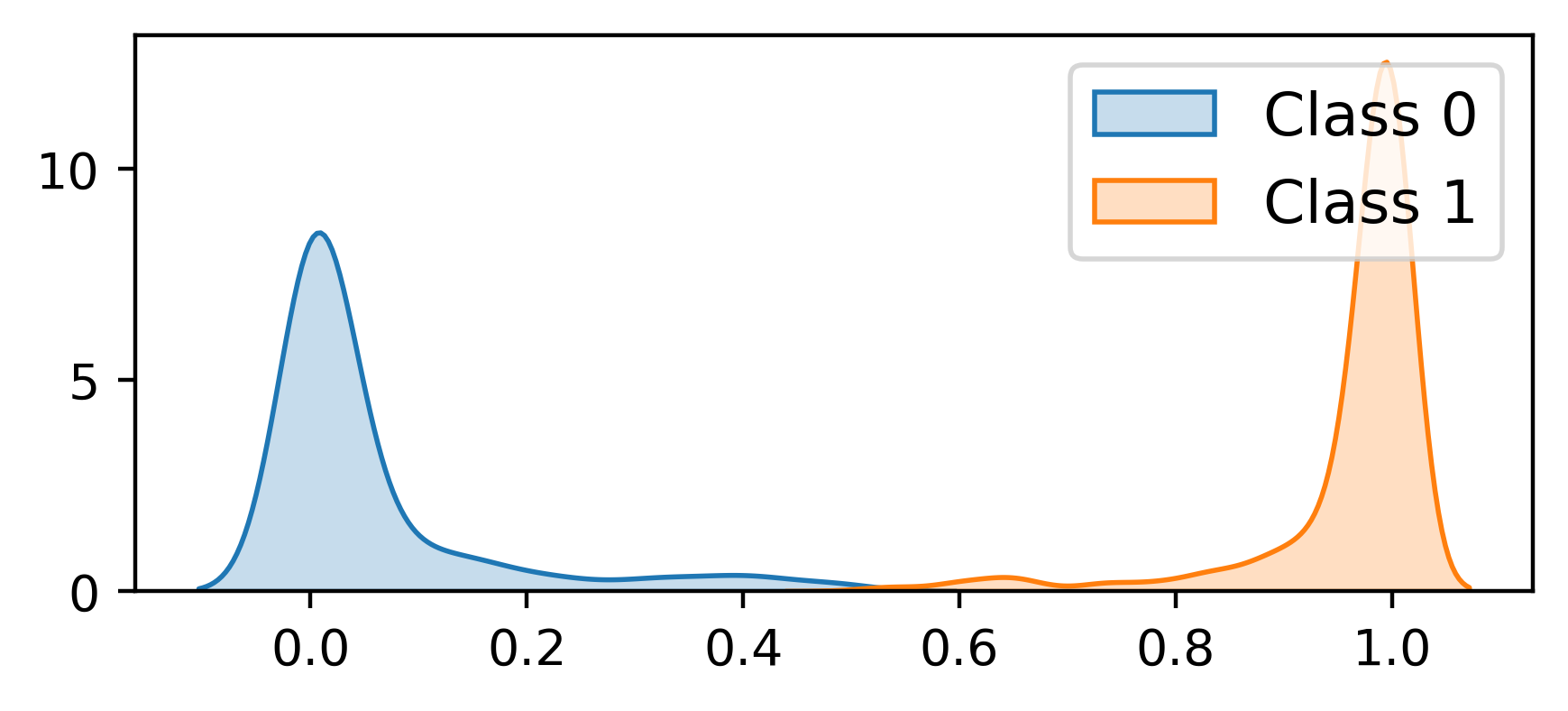}
    \caption{\textbf{IPF} dataset. Comparison of the prediction score distributions. Unlabeled proportion of 40\%. From top to bottom : LabelPropagation, LabelSpreading, FCNN, SSAE.} 
    \label{distrib-ipf-lab}
\end{figure}

Figure \ref{distrib-ipf-lab} displays the distributions of the predicted scores for each class for the unlabelled samples. We can see that our SSAE and the FCNN provide very confident predictions, which is critical in biological applications \cite{johnson-beyond}.

\begin{figure}
    \centering
    \includegraphics[trim={0 0 0 0.8cm},clip,width=0.49\textwidth,height=4.cm]{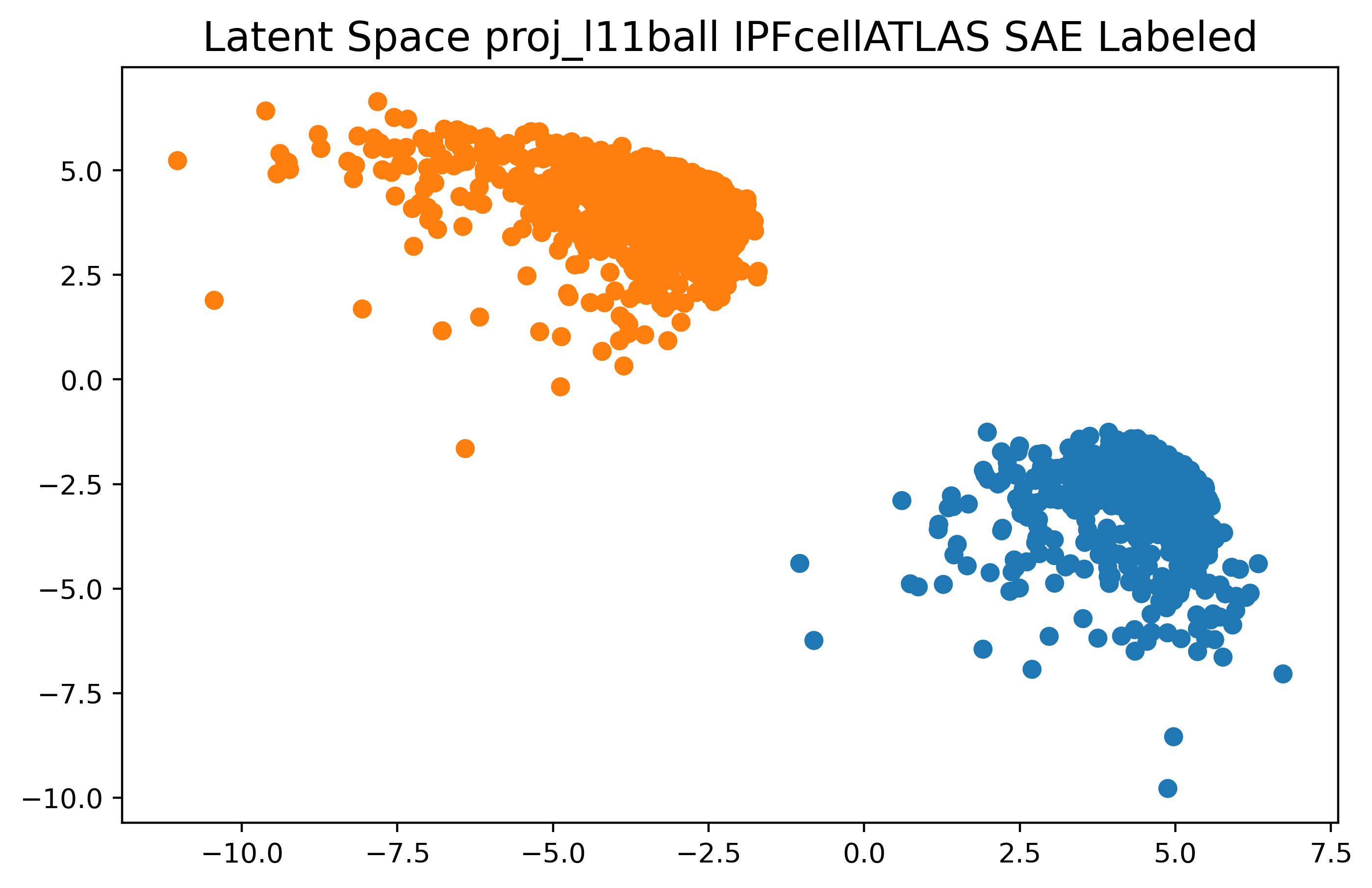}
    \includegraphics[trim={0 0 0 0.8cm},clip,width=0.49\textwidth,height=4.cm]{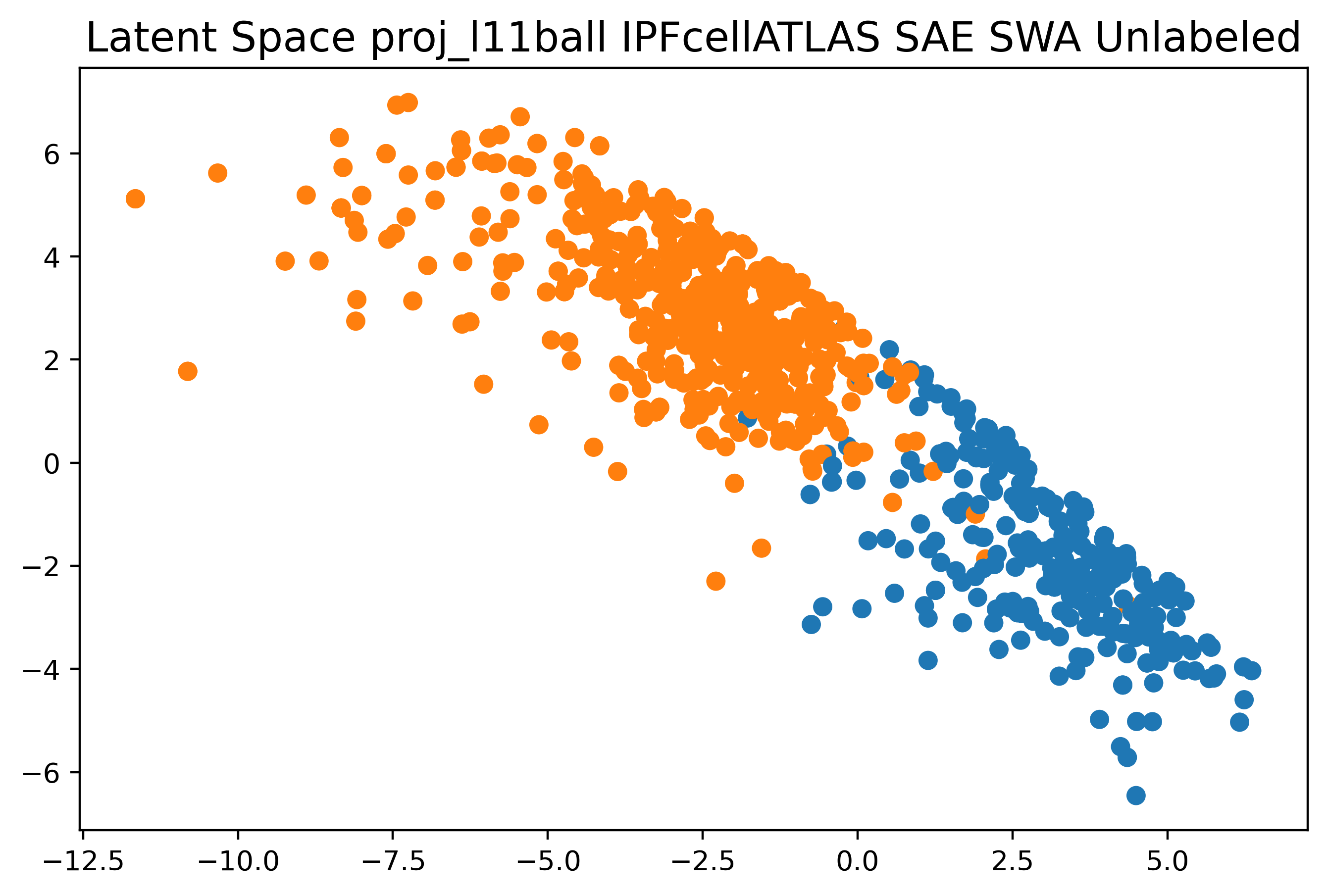}
    \caption{\textbf{IPF} dataset. Samples represented in the latent space of the SSAE. Top: Labeled samples, Bottom: Unlabeled samples.}
    \label{LS-IPF}
\end{figure}

\begin{figure}
    \centering
    \includegraphics[trim={0 0 0 0.8cm},clip,width=0.49\textwidth,height=4.cm]{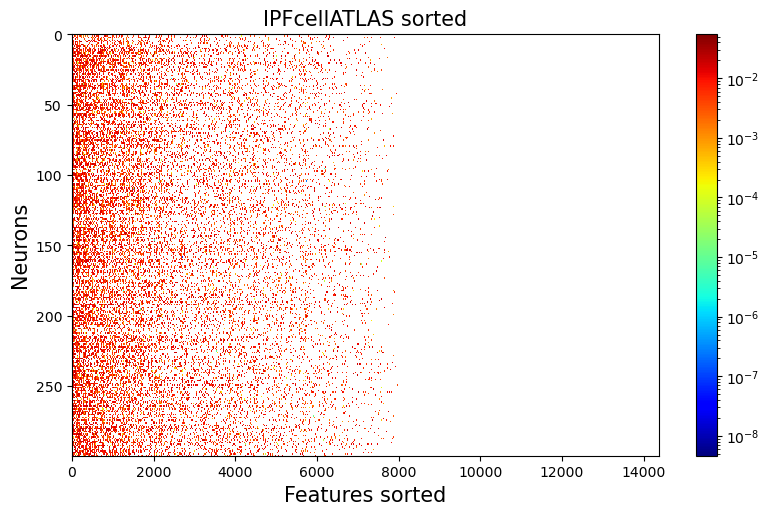}
    \caption{\textbf{IPF} dataset. Weight matrix after projection sorted by feature importance. Logarithmic color scale, zero weights are colored in white.}
    \label{MATRIX-IPF}
\end{figure}

Figure \ref{LS-IPF} shows the latent space of an SSAE trained on the \textbf{IPF} dataset. As on synthetic data, the SSAE learned to perfectly separate the labeled samples, and manages to cluster the two classes on unlabeled data.\\
Figure \ref{MATRIX-IPF} confirms the structured sparsification (see the horizontal lines) abilities of the $\ell_{1,1}$ constraint.\\

The \textbf{LUNG} dataset was provided by Mathe et al. \cite{Lung}. This dataset includes metabolomic data concerning urine samples from $469$ Non-Small Cell Lung Cancer (NSCLC) patients prior to treatment and $536$ control patients. Each sample is described by $2944$ features.

\begin{table}
    \centering
    
    \begin{tabular}{|c|c|c|c|c|c|c|c|}
        \hline
        Lung            & SSAE     & LProp  & LSpread & FCNN  \\
        \hline
        Accuracy $ \%$  & 82.59     &  59.27 &  58.66  & 78.15\\
        \hline
        AUC             &   0.9009  & 0.6569  & 0.6593  & 0.8713\\
       \hline
        F1 score        &   0.8258  & 0.5489 & 0.5399  & 0.7806\\
        \hline
    \end{tabular}
    \caption{\textbf{LUNG} dataset: Mean Metrics over 3 seeds : comparison of LabelPropagation, LabelSpreading, FCNN and SSAE. 40\% of unlabeled data.}
    \label{LUNG-TABLE}
\end{table}


Table \ref{LUNG-TABLE} shows that our SSAE outperforms the classical methods also on this smaller, less balanced dataset. Note that SSAE also outperforms the FCNN by 3\% of AUC and 4\% of both accuracy and F1 Score.

\begin{figure}
    \centering
    \includegraphics[width=0.49\textwidth,height=4.cm]{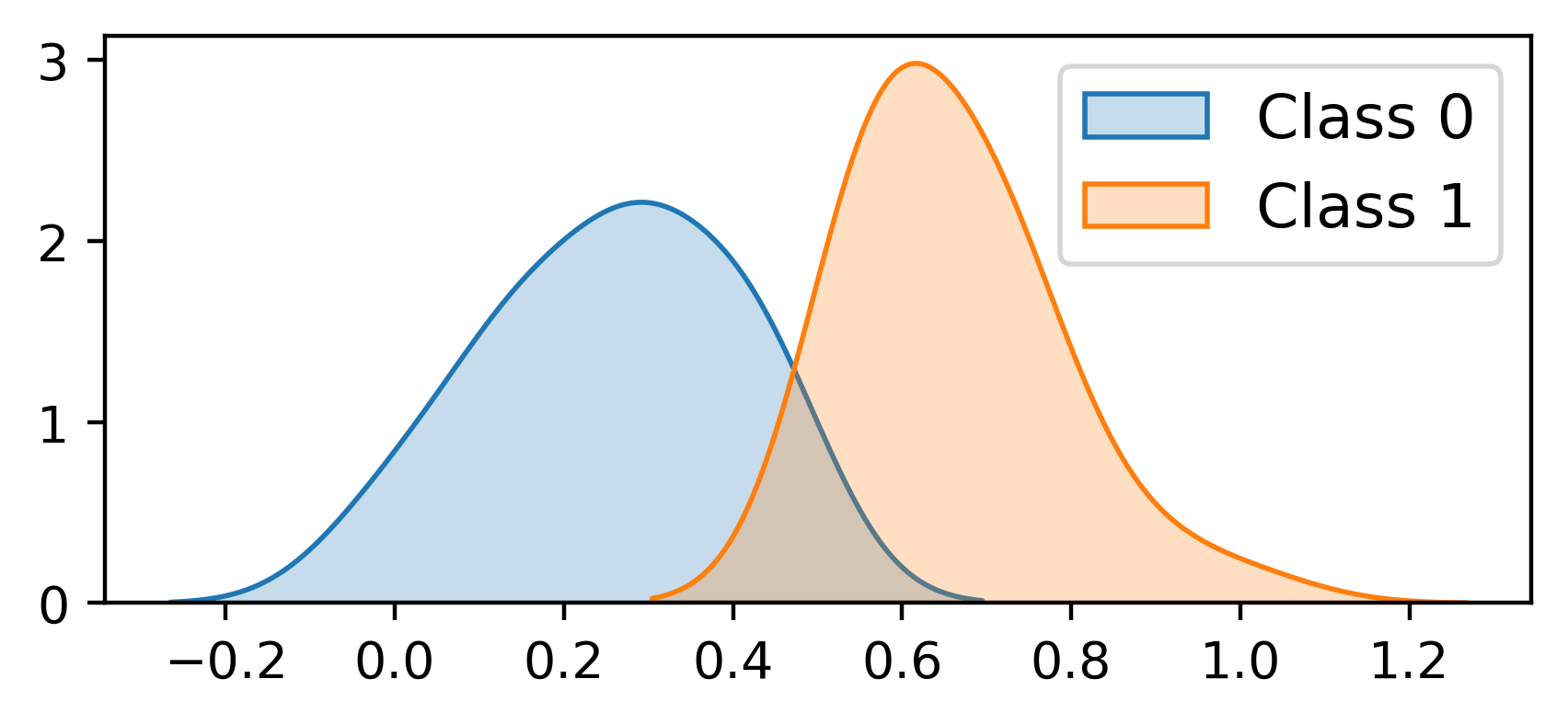}
    \includegraphics[width=0.49\textwidth,height=4.cm]{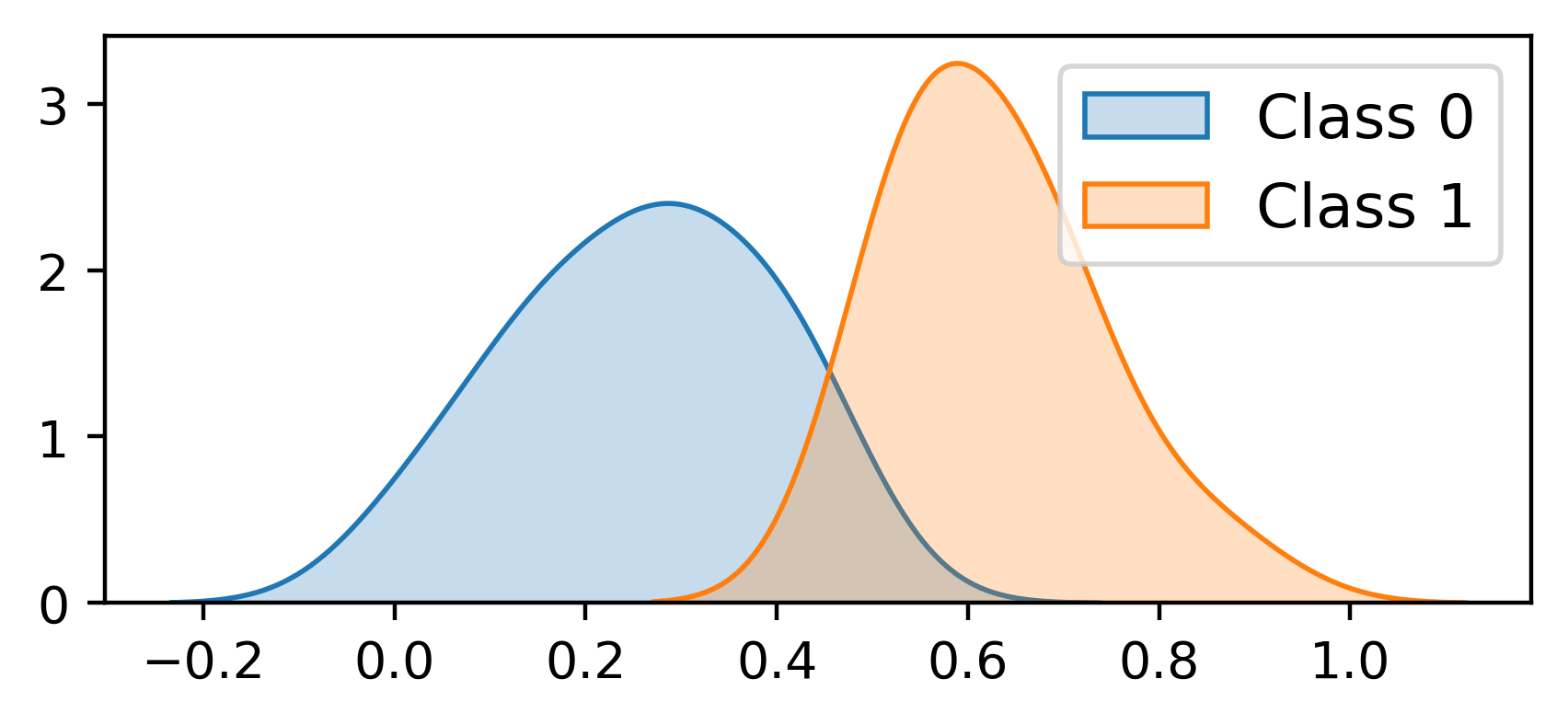}
    \includegraphics[width=0.49\textwidth,height=4.cm]{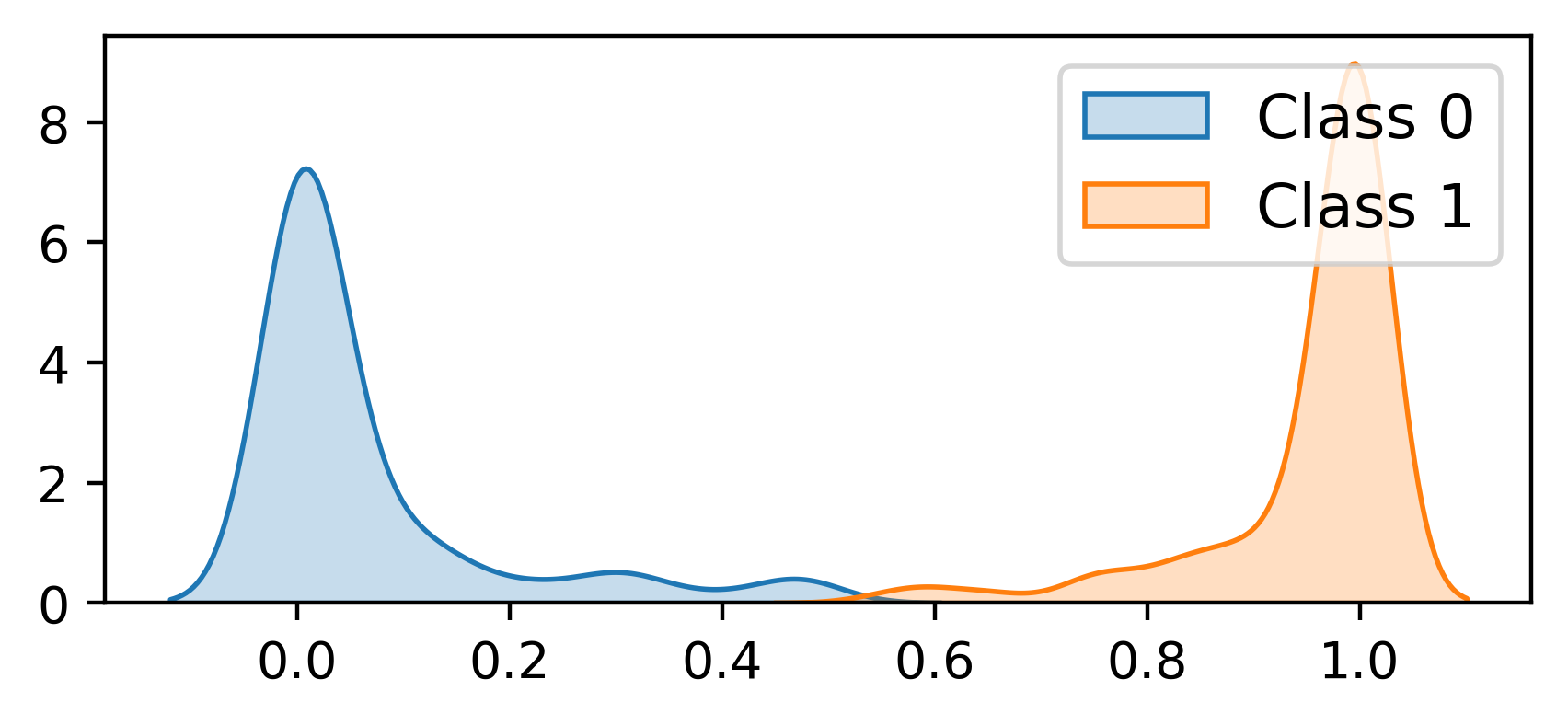}
    \includegraphics[width=0.49\textwidth,height=4.cm]{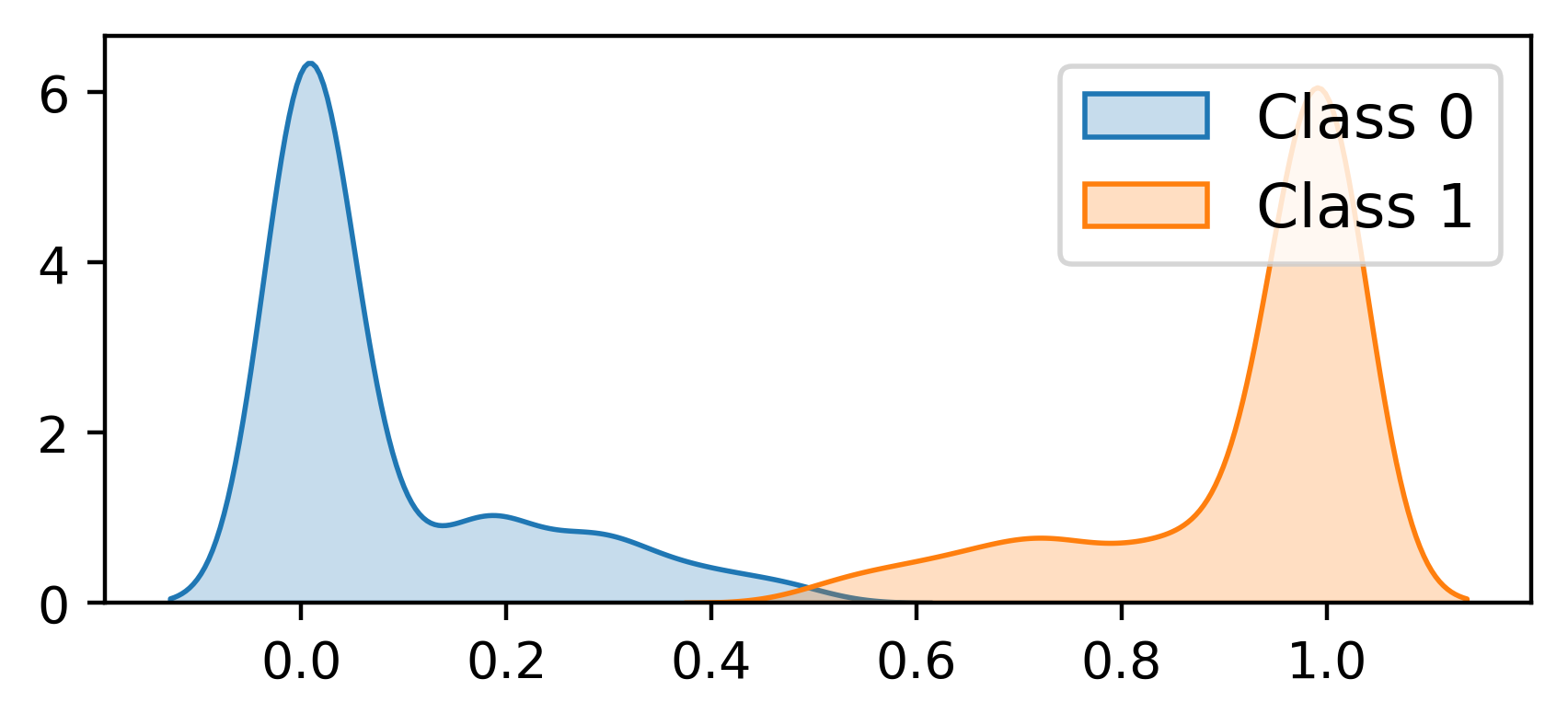}
    \caption{\textbf{LUNG} dataset. Comparison of the prediction score distributions. Unlabeled proportion of 40\%. From top to bottom : LabelPropagation, LabelSpreading, FCNN, SSAE.}
    \label{Distributions-Lung}
\end{figure}
Figure \ref{Distributions-Lung} confirms that the SSAE is able to confidently label the unsupervised samples. We also see that the fully connected neural network will more often lead to low confidence scores for its predictions on this more challenging dataset.

\begin{figure}
    \centering
    \includegraphics[trim={0 0 0 0.8cm},clip,width=0.49\textwidth,height=4.cm]{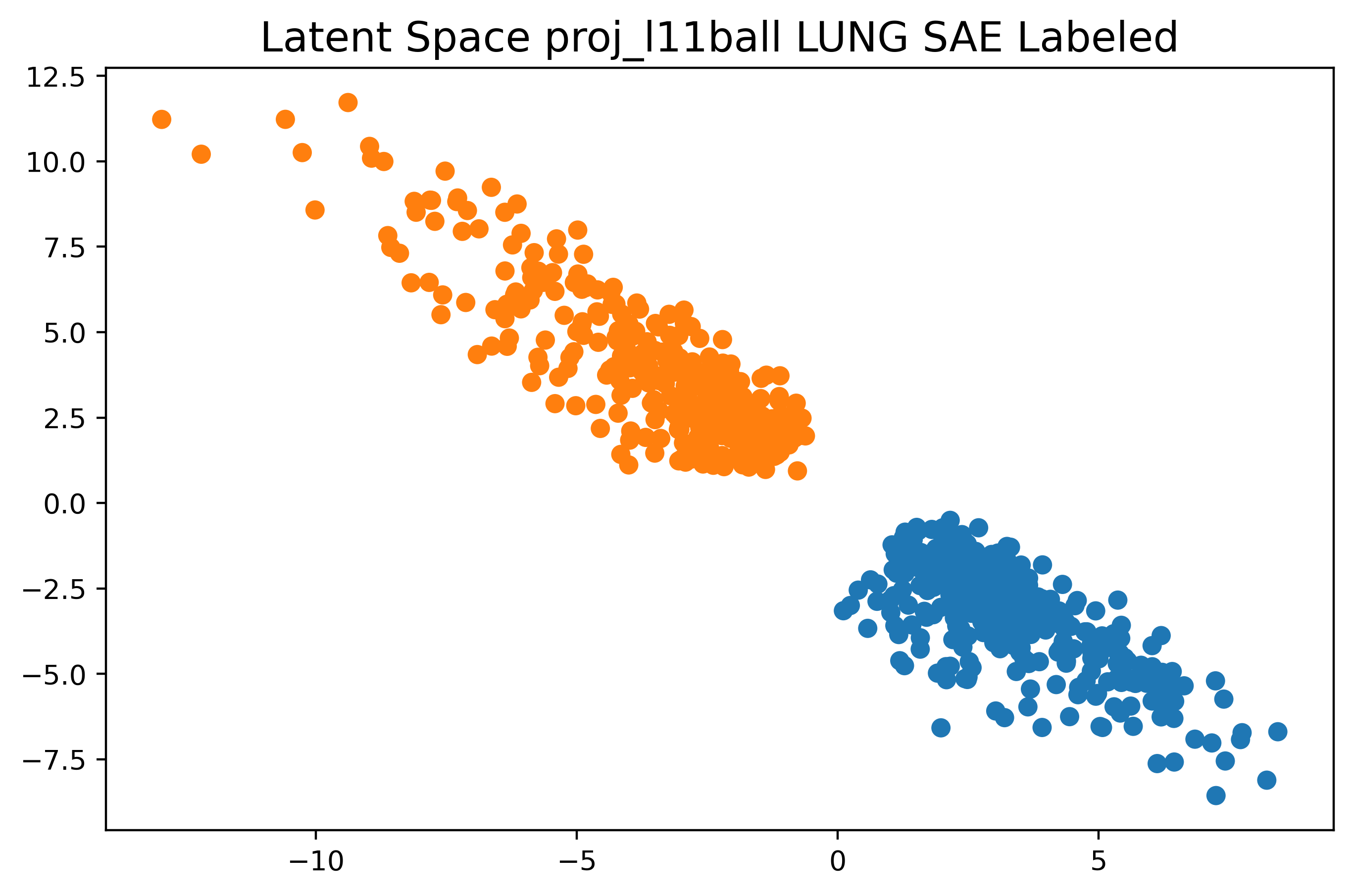}
    \includegraphics[trim={0 0 0 0.8cm},clip,width=0.49\textwidth,height=4.cm]{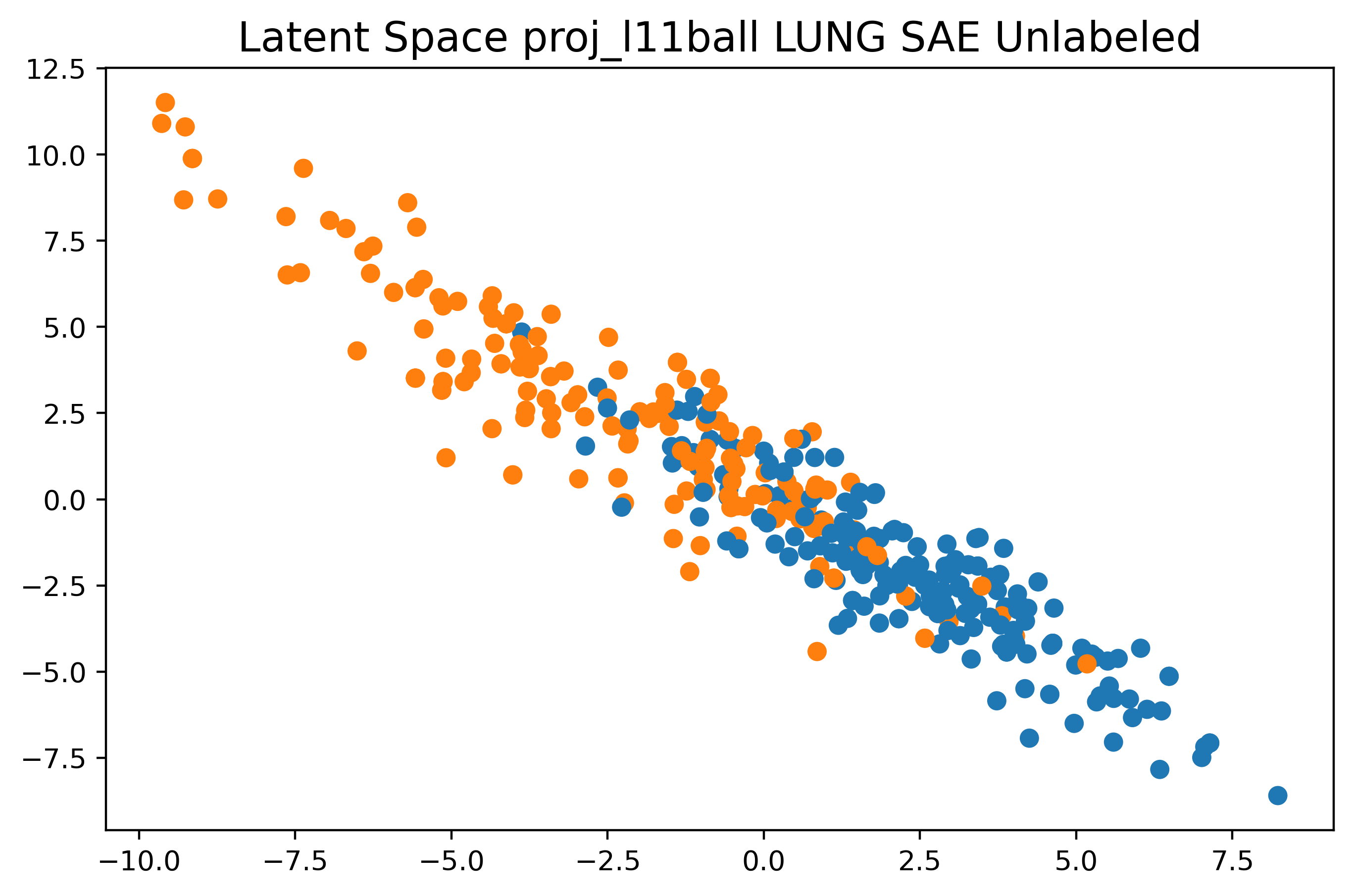}
    \caption{\textbf{LUNG} dataset. Samples represented in the latent space of the SSAE. Top: Labeled samples, Bottom: Unlabeled samples.}
    \label{LS-LUNG}
\end{figure}

\begin{figure}
    \centering
    \includegraphics[trim={0 0 0 0.8cm},clip,width=0.49\textwidth,height=4.cm]{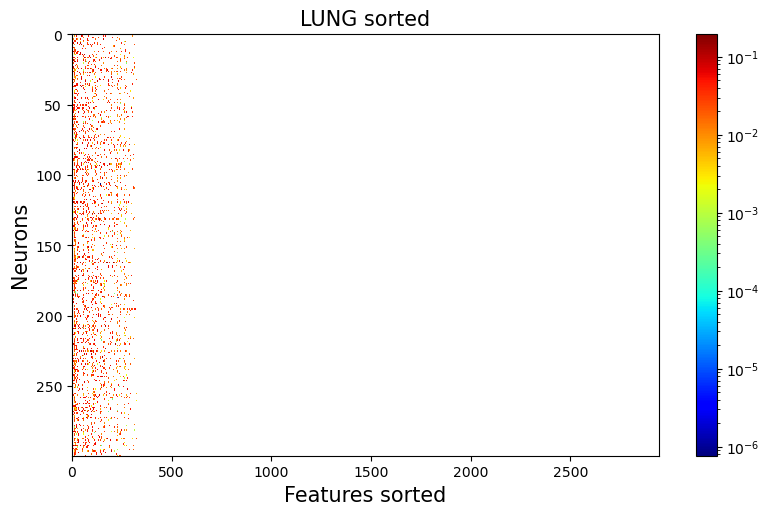}
    \caption{\textbf{LUNG} dataset. Weight matrix after projection sorted by feature importance. Logarithmic color scale, zero weights are colored in white.}
    \label{MATRIX-LUNG}
\end{figure}

Figure \ref{LS-LUNG} shows the latent space of an SSAE trained on the \textbf{LUNG} dataset. This time, while the SSAE still learned to separate the labeled samples, we can see that the lower metrics reported in table \ref{LUNG-TABLE}. translate to more mixed sample clusters, demonstrating the results' interpretability provided by the latent space.

Figure \ref{MATRIX-LUNG} confirms the structured sparsification (see the horizontal lines) abilities of the $\ell_{1,1}$ constraint.




\section{Conclusion}
In this paper we have presented a new framework to solve semi-supervised classification tasks, involving a supervised autoencoder network.  Results show that this approach outperforms classical semi-supervised classification techniques, and provides insightful features for biological applications, such as a latent space, compared to classical fully connected neural networks (FCNN).\\
In a future work, we propose to apply our method to other architectures, such as large Convolutional Neural Networks for image processing.
The algorithm for projecting a matrix on the constraint $\ell_{1,1}$ can be extended to the projection of a tensor on the constraint $\ell_{1,1,1}$ : extension of our method to image processing using convolutional neural networks could thus be straightforward \cite{lossy}.

\section*{Acknowledgments}
The authors thank Marin Truchi and Bernard Mari (IPMC Laboratory) for providing the IPF dataset and Thierry Pourcher (TIRO Laboratory) for providing the Lung dataset.

\bibliographystyle{IEEEtran}
\bibliography{references}

\end{document}